\documentclass{ecai}
\usepackage{latexsym}
\usepackage{amssymb}
\usepackage{times}
\usepackage{soul}
\usepackage{url}
\usepackage[hidelinks]{hyperref}
\usepackage[utf8]{inputenc}
\usepackage[small]{caption}
\usepackage[pdftex]{graphicx}
\usepackage{amsthm}
\usepackage{booktabs}
\usepackage{algorithm}
\usepackage{algorithmic}
\usepackage[switch]{lineno}
\usepackage{bbm}
\usepackage{subcaption}
\usepackage{caption}
\usepackage{tikz}
\usepackage{tabularx}
\usepackage{booktabs, multirow}
\usepackage[english]{babel}
\usepackage{caption}
\usepackage[decisionutilitycolor]{influence-diagrams}
\usepackage{subcaption}
\usepackage{commath}
\usepackage{svg}
\usepackage{makecell}
\usepackage{soul}
\usepackage{pifont}
\usepackage{xspace}
\usepackage{color, colortbl}
\usepackage{float}
\usepackage{booktabs, multirow} 

\usepackage{amsmath}
\usepackage{dsfont}
\usepackage[skip=0.5\baselineskip]{caption}

\usepackage{mwe}
\usepackage{enumitem}
\usepackage{amsthm}
\usepackage{caption}
\usepackage{bbm}
\usepackage{tikz}
\usepackage{tabularx}
\usepackage{adjustbox}

\usepackage[T1]{fontenc}

\usepackage{pgfplots}
\pgfplotsset{compat=1.17}
\definecolor{clrLR}{RGB}{175, 213, 79}
\definecolor{clrDT}{RGB}{225, 160, 79}
\definecolor{clrSVM}{RGB}{225, 0, 79}
\definecolor{clrLGB}{RGB}{207, 52, 118}
\definecolor{clrXGB}{RGB}{32, 137, 227}
\definecolor{clRF}{RGB}{254, 111, 0}
\definecolor{clMLP}{RGB}{13, 6, 134}
\definecolor{clLFERM}{RGB}{134, 197, 189}
\definecolor{clADV}{RGB}{129, 12, 134}
\definecolor{clZafar}{RGB}{252, 119, 147}

\theoremstyle{definition}
\newtheorem{example}{Example}
\newtheorem{definition}{Definition}

\begin{document}

\begin{frontmatter}

\title{Counterfactual Reasoning for Bias Evaluation and Detection in a Fairness under Unawareness setting}
\author[A]{\fnms{Giandomenico}~\snm{Cornacchia}\orcid{0000-0001-5448-9970}\thanks{Corresponding Author. Email: giandomenico.cornacchia@poliba.it}}
\author[A]{\fnms{Vito Walter}~\snm{Anelli}\orcid{0000-0002-5567-4307}}
\author[A]{\fnms{Fedelucio}~\snm{Narducci}\orcid{0000-0002-9255-3256}}
\author[B]{\fnms{Azzurra}~\snm{Ragone}\orcid{0000-0002-3537-7663}}
\author[A]{\fnms{Eugenio}~\snm{Di~Sciascio}\orcid{0000-0002-5484-9945}} 

\address[A]{Polytechnic University of Bari}
\address[B]{Università degli Studi di Bari Aldo Moro}

\begin{abstract}Current AI regulations require discarding sensitive features (e.g., gender, race, religion) in the algorithm's decision-making process to prevent unfair outcomes.
However, even without sensitive features in the training set,  algorithms can persist in discrimination.
Indeed, when sensitive features are omitted (\textit{fairness under unawareness}), they could be inferred through non-linear relations with the so-called proxy features.
In this work, we propose a way to reveal the potential hidden bias of a machine learning model that can persist even when sensitive features are discarded.
This study shows that it is possible to unveil whether the black-box predictor is still biased by exploiting counterfactual reasoning.
In detail, when the predictor provides a negative classification outcome, our approach first builds counterfactual examples for a discriminated user category to obtain a positive outcome. Then, the same counterfactual samples feed an external classifier (that targets a sensitive feature) that reveals if the modifications to the user characteristics needed for a positive outcome moved the individual to the non-discriminated group. When this occurs, it could be a warning sign for discriminatory behavior in the decision process. Furthermore, we leverage the deviation of counterfactuals from the original sample to determine which features are proxies of specific sensitive information.
Our experiments show that, even if the model is trained without sensitive features, it often suffers discriminatory biases. 
\end{abstract}

\end{frontmatter}

\section{Introduction}
\label{sec:introduction}

The use of AI systems that deal with life-changing tasks is increasing daily, raising  social concerns. 
One of the most worrisome is the \textit{discrimination}
of minority groups 
or individuals.
As an example, in the financial domain, the decision to approve or deny credit has been regulated with precise and detailed regulatory compliance requirements (i.e., Equal Credit Opportunity Act
, Federal Fair Lending Act
, and Consumer Credit Directive for EU Community).
These rules aim to prevent discrimination in human decision-making processes. However, they do not fit scenarios involving Machine Learning (ML) or, more broadly, Artificial Intelligence (AI) systems.
In the wake of the GDPR
(i.e., a regulatory oversight to protect personal information), the EU Commission has issued ``Ethics Guidelines for Trustworthy AI'' and, more recently, ``The Proposal for Harmonized Rule on AI'' to prevent irresponsible uses of such technology.
Several studies have shown that deploying AI systems without considering ethical concerns might encourage discrimination, although system designers train their models without discriminatory intent~\cite{DBLP:conf/kdd/Corbett-DaviesP17,doi:10.1126/sciadv.aao5580,doi:10.1126/science.187.4175.398,CORNACCHIA2023103224, cornacchia2023counterfactual}.
Moreover, omitting sensitive features to train AI models does not guarantee the absence of biases in the outcomes~\cite{agarwal2021responsible,DBLP:journals/corr/abs-2009-06251,article}.
Indeed, there may exist \textit{proxy features}, which behave~as implicit sensitive~feature (e.g., I can infer the gender from the individual's~habits).
This study proposes an approach for detecting biased decisions in a realistic scenario where sensitive features are omitted and an unknown number of proxy features may be present.
Our designed pipeline is composed of three main modules. The \textbf{Decision Maker} is the first module, which wraps the critical classifier to analyze, and is trained without using the sensitive features (e.g., it decides to grant or not a loan).
The second module trains a different classifier, named \textbf{Sensitive-Feature Classifier}, on the same features to predict the sensitive characteristics (e.g., it classifies if the individual involved in the previous decision is a member of the protected or non protected group). The third module, the \textbf{Counterfactual generator}, calculates the minimal counterfactual samples by modifying the values of non-sensitive features to obtain the desired outcome on the Decision maker (e.g., loan approved).
Eventually, the sensitive feature predictor classifies the generated counterfactual samples to check whether the samples do still belong to the original sensitive class (e.g., gender female).
If this does not occur (e.g., the new counterfactual sample obtaining the loan is classified now as male, opposite to the original class), the outcome predictor is biased, and its unfairness can be quantified.

Let us provide a real-life example to explain our model:
\begin{example}Anna is a young researcher who wants to buy a house and decides to apply for a loan. She got a permanent contract two months ago, has a good income, and likes cinema, open-air sports, and visiting museums.
The bank AI system analyzed Anna's financial profile, non-sensitive information, bank account transactions, and income data and decided to deny the loan. At this point, our system comes into play to discover whether the decision is affected by bias. It acquires all of Anna's non-sensitive data exploited by the AI bank system and starts to generate counterfactual samples. Consequently, Anna's profile is slightly changed regarding employment-contract duration, monthly income, and recurrent transactions until the new counterfactual profile gets a favorable decision by the AI system and the loan is approved. Anna's new profile is now used to feed the sensitive-feature classifier. The sensitive classifier decides that Anna's original profile belongs to the female class, but the new counterfactual profile belongs to the male class. This demonstrates decision bias since, even though the system does not exploit sensitive features and does not know Anna's gender, it classifies Anna's counterfactual profile (who gets the loan) as belonging to the (privileged) male class.
\end{example}

%

To summarize, we propose an approach for detecting bias in machine learning models making use of counterfactual reasoning, even when those models are trained without sensitive features, i.e., the setting known as \textit{Fairness Under Unawareness}~\cite{DBLP:conf/fat/ChenKMSU19,mehrabi2021survey}. 
In detail, our work aims to answer the following research questions:
\begin{itemize}

    \item \textbf{RQ1:} Is there a method for determining whether a dataset contains proxy features or not?
    \item \textbf{RQ2:} Does the Fairness Under Unawareness setting ensure that decision biases are avoided? 
    \item \textbf{RQ3:} Is counterfactual reasoning effective for discovering decision biases?
    \item \textbf{RQ4:} Is it possible to define a strategy for identifying the proxy features?
\end{itemize}
The remainder of the paper is organized as follows: Section~\ref{sec:bibliography} provides an overview of the most relevant research in the fields of fairness and counterfactual reasoning, Section~\ref{sec:preliminaries} introduces the preliminaries, Section~\ref{sec:approach} depicts our methodology, Section~\ref{sec:experiments} describes the experimental evaluation, Section~\ref{sec:results} discusses the results, and the conclusion and future work are drawn in Section~\ref{sec:conclusion}. 
\noindent \textbf{Code is available at:} 
\url{https://github.com/giandos200/ECAI23}.

\section{Related Work}
\label{sec:bibliography}
The section illustrates the most significant studies on \textit{fairness under unawareness} and \textit{counterfactual reasoning}~research.

\paragraph{Fairness, Fairness Under Unawareness, and Proxy Features.}
\label{sec:fairness_under_unawareness}
In machine learning research, fairness is a well-studied topic with a considerable body of knowledge to draw from~\cite{DBLP:journals/ipm/AshokanH21,DBLP:conf/cikm/ZhuHC18,DBLP:conf/kdd/PedreschiRT08,crenshaw2013mapping}.
In fact, due to the critical impact of decision-making on people's lives (e.g., Financial Services, Education, and Health), the use of sensitive characteristics is strictly prohibited.
While designing the decision-making algorithm not to leverage sensitive information is simple, assuring the same accuracy as before and demonstrating that the predictor is unbiased, despite the absence of such sensitive characteristics, is another matter~\cite{Chen2018d,cornacchia2023general,DBLP:conf/recsys/CornacchiaNR21}. 

The system may infer sensitive features from variables, i.e. proxy variables, that nonetheless represent protected characteristics, even if the user does not explicitly give sensitive information.
The models that infer sensitive features from proxy variables are known as ``probabilistic proxy models"~\cite{DBLP:conf/fat/ChenKMSU19,ConsFinancial}.
The majority of the approaches for identifying proxy features proposed in the literature rely on strategies for determining multicollinearity between variables~\cite{agarwal2021responsible,yeom2018hunting}. 
The relationships, however, may not be linear. Cosine similarity and mutual information are the most commonly employed methods in this scenario, according to Agarwal and Mishra~\cite{agarwal2021responsible}.
Chen et al.~\cite{DBLP:conf/fat/ChenKMSU19} studied the relationship between proxy features, sensitive variables (i.e., geolocation and race), and classifiers threshold. 
Fabris et al.~\cite{DBLP:journals/corr/abs-2109-08549} use a quantification approach to measure group fairness in an unawareness setting. 

\paragraph{Counterfactual Reasoning.}
\label{sec:counterfactual_reasoning}
Counterfactual reasoning is a flourishing field in artificial intelligence research~\cite{DBLP:journals/ai/Ginsberg86,DBLP:journals/ai/Miller19,DBLP:conf/recsys/CornacchiaNR21}. This research was initially born to investigate causal links~\cite{DBLP:conf/ecai/Pearl94}, and today it can count on several contributions~\cite{DBLP:conf/context/Ferrario01}.
Most of them define and employ counterfactuals as helpful tools to explain the decisions taken by modern decision support systems. 
The study that takes the first steps from the same motivations was conducted by Kusner et al.~\cite{DBLP:conf/nips/KusnerLRS17} which proposed a Counterfactual metric. The metric exploits causal inference to assess fairness at an individual level by requiring that a sensitive attribute not be the cause of a change in a prediction. In the same direction, Wu et al.~\cite{DBLP:conf/ijcai/Wu0W19} propose a bounded linear constrained optimization of countefactual faireness addressing the limitation of Kusner et al.~\cite{DBLP:conf/nips/KusnerLRS17} work.
For what concerns the risk assessment domain, Mishler et al.~\cite{DBLP:conf/fat/MishlerKC21} put forward a similar working hypothesis. 
In detail, they propose a counterfactual equalized odds ratio criterion to train predictors operating in the post-processing phase.
They extend and adapt previous post-processing approaches~\cite{DBLP:conf/nips/HardtPNS16} to the counterfactual setting.

\paragraph{Our Contribution} Differently from the majority of the recent studies, our investigation aims to leverage a counterfactual generation tool to reveal the presence of implicit biases in a decision support system seen as a black box.
Interestingly, this motivation is similar to Bottou et al.~\cite{bottou2013counterfactual}.
Indeed, we both aim to answer the question: "How would the system have decided if we had replaced some user characteristics?". The paper focuses on exploiting possible system behavior changes in real-life applications of minimal intervention. In contrast, our contribution focuses on the same work philosophy but on exploiting possible unethical consequences of machine learning systems.
Beyond this commonality, the two studies differ significantly as Bottou et al.~\cite{bottou2013counterfactual} focus on measuring the fidelity level of the system.

\section{Preliminaries }
\label{sec:preliminaries}
This section introduces the notation adopted hereinafter.

\noindent\textbf{Data points:} We assume the dataset $\mathcal{D}$ is an $m$-dimensional space containing $n$ non-sensitive features, $l$ sensitive features, and a target attribute. In other words, we have $\mathcal{D} \subseteq \mathbb{R}^{m}$, with $m = n+l+1$.\footnote{Without loss of generality, we assume that categorical features can always be transformed into features in $\mathbb{R}$ via one-hot-encoding.} A data point $d\in\mathcal{D}$ is then represented as $d=\langle \mathbf{x}, \mathbf{s}, y \rangle$, with $\mathbf{x} = \langle x_{1}, x_{2},..., x_{n}\rangle$ representing the sub-vector of non-sensitive features, $\mathbf{s} = \langle s_{1}, s_{2},..., s_{l}\rangle$ the sub-vector of sensitive features and $y$ being a binary target feature. 
Given a vector of sensitive festures, $\forall s_i \in \mathbf{s}$,
$s_i^{-}$ refers to the \textit{unprivileged} group and $s_i^{+}$ to the \textit{privileged} group of the $i$-th sensitive feature. 
\\
\noindent\textbf{Proxy Features:} 
$\mathbf{p}_i \subseteq \mathbf{x}$ is a subset of features for a sensitive feature $s_i$ such that there exists a function $h(\cdot)$ which can estimate $s_i$ (i.e., $h(\mathbf{p}_i) = s_i$).
\\
\noindent\textbf{Target Labels:} Given a target feature $y \in \{0,1\}$, $y=1$ is the positive outcome and $y=0$ is the negative one. 
\\
\noindent\textbf{Outcome Prediction:}
$\hat{y} \in \{0,1\}$ represents the prediction for $\mathbf{x} \subset d$ estimated by $f(\cdot)$, a function such that $f(\mathbf{x}) = \hat{y}$.
\\
\noindent\textbf{Sensitive Feature Prediction:}
$\hat{s}_i \in \{-,+\}$ represents the prediction of the $i$-th sensitive feature for a given data point estimated by $f_{s_i}(\cdot)$, a function s.t. $f_{s_i}(\mathbf{x}) = \hat{s}_i$.
\\
\noindent\textbf{Counterfactual samples:}
Given a vector $\mathbf{x}$ and a perturbation $\mathbf{\epsilon}=\langle\epsilon_{1}, \epsilon_{2},..., \epsilon_{n}\rangle$, we say that a vector $ \mathbf{c}_\mathbf{x}=\langle c_{x_1}, c_{x_2},..., c_{x_n}\rangle= \mathbf{x} + \mathbf{\epsilon}$ is a counterfactual (CF) of $\mathbf{x}$ if $f(\mathbf{c}_\mathbf{x}) = 1- f(\mathbf{x})= 1- \hat{y}$. We use the set $\mathcal{C}_\mathbf{x}$, with $\lvert \mathcal{C}_\mathbf{x} \rvert = k$, to denote the set of possible \textbf{counterfactual samples} for $\mathbf{x}$. A function $g(\mathbf{x})$ is used to compute $k$ counterfactuals for $\mathbf{x}$.
\\
For simplicity, we denote $f(\cdot)$, $f_{s_i}(\cdot)$, and $g(\mathbf{x})$ as the \textbf{Decision Maker}, the \textbf{Sensitive-Feature Classifier}, and the \textbf{Counterfactual Generator} respectively.

\section{Methodology}
\label{sec:approach}
\begin{figure*}[!t]
    \centering
    \includegraphics[width=0.8\textwidth]{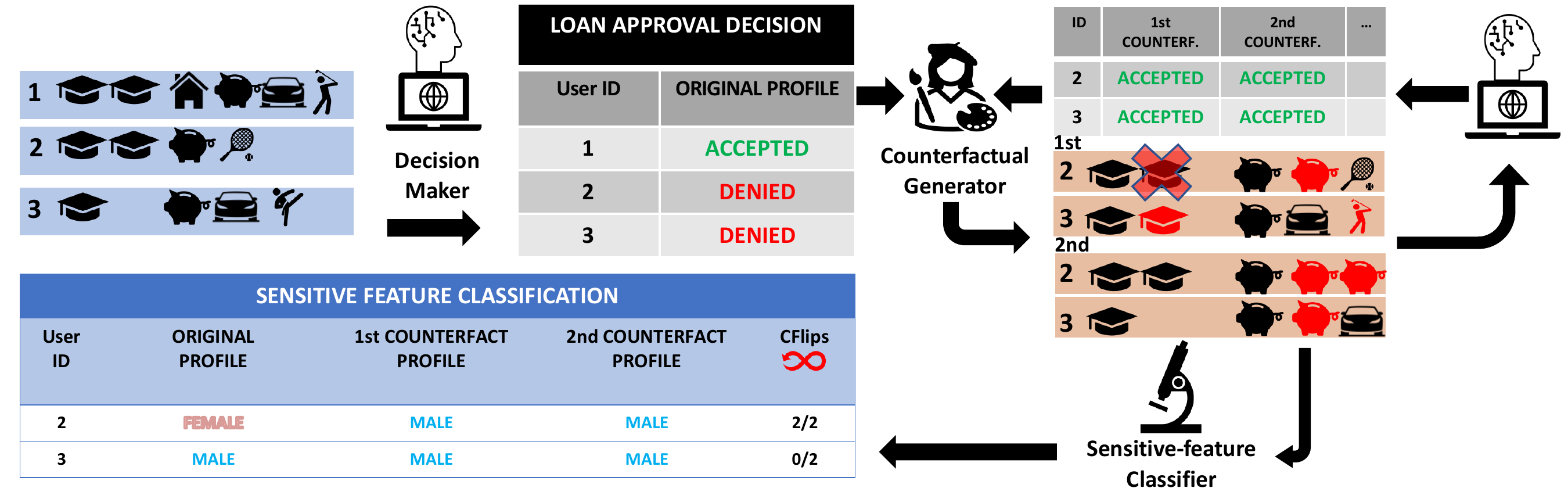}
    \caption{An example of a loan-approval decision  process analyzed through our model. From left to right, we have the decision made on the original user profiles, the counterfactual generation for users with loan denied, the sensitive-feature classification of the original profile, and counterfactual profiles with the decision changed. For user~2, both counterfactual profiles change the sensitive-feature category.}
    \label{fig:CFgen}
\end{figure*}

The \textit{fairness under unawareness} setting (see Section~\ref{sec:fairness_under_unawareness}) poses several challenges to the identification of discriminatory behaviors performed by  intelligent systems. 
Proxy traits can be non-linearly associated with sensitive ones, making typical statistical procedures ineffective. 
Figure \ref{fig:CFgen} depicts the principal components of our model, namely the \textit{Decision-Maker}, the \textit{Counterfactual Generator}, and the \textit{Sensitive-Feature Classifier}, as well as the flow of our pipeline.
As a relevant case study, we refer to the financial domain, considering the tasks of predicting loan-repayment default. However, focusing on this specific domain does not compromise the model generality.
\begin{example}\label{ex:exmp2}
Let us imagine that some users with certain characteristics apply for a loan (see Figure \ref{fig:CFgen}). The \textit{Decision Maker} analyzes their requests and computes a positive or negative decision. If a request is denied (e.g., for users 2 and 3), the \textit{counterfactual generator} starts to produce a series of counterfactuals until
 it gets a positive decision ($\hat{y}=1$, loan accepted). In the example, only two counterfactuals for users 2 and 3 are generated. Once the decision has been changed, the \textit{Sensitive-Feature Classifier} analyzes the original characteristics of users' 2 and 3 profiles and the newly generated counterfactuals to assess how many counterfactuals changed the sensitive-feature \textit{gender}. User 2 was originally classified as female and, then, as male for both counterfactuals (profile with counterfactual changes for getting loan approval). For user 3, this does not happen, the gender classification is the same before and after the counterfactual changes (loan approved).
\end{example}


To define a discrimination score of a given decision model, we propose a metric that we call \textit{Counterfactual Flips} providing a snapshot of the discriminatory behavior the model might put in place.
\begin{definition}[\textit{Counterfactual Flips}] \textit{Given a sample $\mathbf{x}$ belonging to a demographic group $s$ whose model output is denoted as $f(\mathbf{x})$, generated a set $\mathcal{C}_\mathbf{x}$ of $k$ counterfactuals with a desired $y^{*}$ outcome $f(\mathbf{c}_\mathbf{x}^i) = y^{*} \quad \forall \mathbf{c}_\mathbf{x}^i \in \mathcal{C}_\mathbf{x}$, the Counterfactual Flip indicates the percentage of counterfactual samples belonging to another demographic group (i.e., $f_s(\mathbf{c}_\mathbf{x}^i) \neq f_s(\mathbf{x})$, with $f_s(\mathbf{x})=s$).}
\begin{equation}\label{eq:CFlips}
\footnotesize
\mathrm{CFlips}(\mathbf{x},\mathcal{C}_{\mathbf{x}},f_s(\cdot)) = \frac{\sum_{i=1}^k(\mathbbm{1}(\mathbf{c}_\mathbf{x}^i))}{k} 
\end{equation}
where the function $\mathbbm{1}(\mathbf{c}_\mathbf{x}^i)$ correspond to:
\begin{equation}
\footnotesize
     \mathbbm{1}(\mathbf{c}_\mathbf{x}^i) = \begin{cases}
1 &\text{if }  f_s(\mathbf{c}_\mathbf{x}^i) \neq f_s(\mathbf{x}) \neq s \\
0 &\text{if }  f_s(\mathbf{c}_\mathbf{x}^i) = f_s(\mathbf{x}) = s
\end{cases}
\end{equation}
and it measures if a counterfactual sample is linked with a \textit{Flip}, i.e. a change, for the sensitive characteristic.
\end{definition}
The bigger the CFlips value is, the stronger the biases and the discrimination the model suffers from.
In Example~\ref{ex:exmp2}, $CFlips=1$ for user 2, and $CFlips=0$ for user 3, thus the sensitive classification changed 2 out of 2 CF samples for user 2 and 0 out of 2 CF samples for user 3, unveiling implicit bias in the \textit{Decision Maker} in favor of male characteristics.
Therefore, we can introduce the proposed \textit{Counterfactual Fair Flips} desiderata. 

\begin{definition}[Counterfactual Fair Flips]\label{not:teorem1}
-
A binary classifier shows Counterfactual Fair Flips if the probability of generating Counterfactual samples belonging to a different demographic group (privileged vs unprivileged) is the same:

\begin{equation}\label{eq:theorem}
\footnotesize
    \mathds{P}(f_s(\mathcal{C}_{\mathcal{D}\vert_{s^-}}) \neq s^- \mid f(\mathcal{C}_{\mathcal{D}\vert_{s^-}}))=\mathds{P}(f_s(\mathcal{C}_{\mathcal{D}\vert_{s^+}})\neq s^+ \mid f(\mathcal{C}_{\mathcal{D}\vert_{s^+}})) 
\end{equation}
{\normalfont which implies the complement}:
\begin{equation}\label{eq:theorem2}
\footnotesize
    \mathds{P}(f_s(\mathcal{C}_{\mathcal{D}\vert_{s^-}}) = s^- \mid f(\mathcal{C}_{\mathcal{D}\vert_{s^-}}))=\mathds{P}(f_s(\mathcal{C}_{\mathcal{D}\vert_{s^+}}) = s^+ \mid f(\mathcal{C}_{\mathcal{D}\vert_{s^+}})) 
\end{equation}

\end{definition}


In our work, we only take into account samples negatively predicted by the \textit{Decision Maker} (i.e., $f(\mathbf{x})=0$), that we denote as $\mathcal{X}^- \subseteq \mathcal{D}$, as we are interested in quantifying the discrimination of the minority group (e.g. women) in the process to achieving a positive counterfactual result (i.e., $f(\mathbf{c}_\mathbf{x})=1 \land f_s(\mathbf{c}_\mathbf{x}) \neq s$).
For the set of samples $\mathcal{X}^-$,
the metric in Equation~\ref{eq:CFlips} can be generalized to the \textit{privileged} and \textit{unprivileged} group (Equation~\ref{eq:CFlipsPriv} is restricted to the \textit{privileged} samples negatively predicted, while Equation~\ref{eq:CFlipsUnpriv} is restricted to the \textit{unprivileged} samples negatively~predicted).
\begin{equation}\label{eq:CFlipsPriv}
\footnotesize
\text{\parbox{5.0em}{\textbf{\textit{Privileged}}}} \begin{cases} \mathrm{CFlips} =  \frac{\sum_{i=1}^{n}\mathrm{CFlips}(\mathbf{x_i},\mathcal{C}_{\mathbf{x_i}},f_s(\cdot)) }{\lvert \mathcal{X}\vert_{s^+}^{-}\rvert} \quad \text{with } \mathbf{x} \in \mathcal{X}\vert_{s^+}^{-}\end{cases}
\end{equation}
\begin{equation}\label{eq:CFlipsUnpriv}
    \footnotesize
    \text{\parbox{5.5em}{\textbf{\textit{Unprivileged}}}} \begin{cases} \mathrm{CFlips} =  \frac{\sum_{i=1}^{n}\mathrm{CFlips}(\mathbf{x_i},\mathcal{C}_{\mathbf{x_i}},f_s(\cdot)) }{\lvert \mathcal{X}\vert_{s^-}^{-}\rvert} \quad \text{with } \mathbf{x} \in \mathcal{X}\vert_{s^-}^{-}\end{cases}
\end{equation}

Definition~\ref{not:teorem1} requires that the quantities of counterfactuals flipped in equations 5 and 6 must be equal. Thus, we are interested in the difference between the result of the two equations, i.e. $\Delta\mathrm{CFlips}$, being close to zero.

\section{Experimental Design}
\label{sec:experiments}
This section provides the description of our experimental settings, designed to answer the four RQs outlined in Section~\ref{sec:introduction}. Extended details are in Appendix~\ref{sec:AppendixSetup}.
\begin{table}
    \scriptsize
    \centering
    \caption{Adult, Adult-debiased, Crime, and German datasets' characteristics.
    $\lvert n\rvert^{*} $ correspond to the number of non-sensitive features, and $\lvert n\rvert^{\dagger}$ to the number of feature w/o the sensitive one (i.e., \textit{gender} and \textit{crime}).}
    \label{tab:datasetSplit}\label{tab:datasetAdultdebiased}\label{tab:datasetAdult}\label{tab:datasetCrime}\label{tab:datasetGerman}
\setlength{\tabcolsep}{6.5pt}
\begin{tabular}{llrrll}
\hline\toprule
Dataset&Split&$\lvert \mathcal{D} \rvert$ &$\lvert n\rvert^{\dagger}$ &Target ($Y$) &\textbf{$Y=1$} \\\hline\midrule
Adult&\textbf{Train} &40699 &13 &income &$\geq \$50,000 $\\
&\textbf{Test} &4523 &13 &income &$\geq \$50,000$ \\\midrule
Adult-debiased&\textbf{Train} &40699 &6 &income &$\geq \$50,000 $\\
&\textbf{Test} &4523 &6 &income &$\geq \$50,000$ \\ \midrule
Crime&\textbf{Train} &1794 &98 &Violent State &$<$\textit{median} \\ 
&\textbf{Test} &200 &98 &Violent State &$<$\textit{median} \\\midrule
German&\textbf{Train} &900 &17 &credit score &Good \\
&\textbf{Test} &100 &17 &credit score &Good \\
\hline\bottomrule\hline
\end{tabular}
\vspace{-1em}
\end{table}
\paragraph{Datasets.} Experiments have been conducted on three state-of-the-art (SOTA) datasets, used as benchmarks in several  works~\cite{DBLP:conf/iclr/BalunovicRV22,DBLP:conf/nips/DoniniOBSP18,DBLP:conf/kdd/PedreschiRT08,Das33}. These  are: Adult~\cite{kohavi_becker_1996}, a real-world dataset used for income prediction,\footnote{\underline{Adult}: \url{https://archive.ics.uci.edu/ml/datasets/adult}} German~\cite{german_hoffman}, a real-world dataset for default prediction,\footnote{\underline{German}:~\url{https://archive.ics.uci.edu/ml/datasets/statlog+(german+credit+data)}} and Crime~\cite{crime_data}, a real-world Census dataset for violent state prediction\footnote{\underline{Crime}:~\url{https://archive.ics.uci.edu/ml/datasets/US+Census+Data+(1990)}} (i.e., a state is violent if the number of crime in a state is higher with respect to the median($| \mathcal{C}_\mathbf{x} |$) of all the states). We point to the reader attention that the privileged and unprivileged have been chosen based on the ex-ante Statistical Parity in Table~\ref{tab:SFdataDisrib} (i.e., if positive, the first group is the \textit{privileged}, otherwise the second).
For Adult and German, the sensitive attribute we considered is \textit{gender}, with \textit{male} and \textit{female} corresponding to the \textit{privileged} and \textit{unprivileged}~group respectively. For the Crime dataset, the sensitive attribute we considered is the \textit{race} that indicates the race with the largest number of crimes committed in a specific state. As a second sensitive feature, for Adult, we chose \textit{maritalStatus}, with \textit{married} and \textit{not married} as \textit{privileged} and \textit{unprivileged}, and for German, we chose age as $>25$ \textit{years} and $<=25$ \textit{years} as \textit{privileged} and \textit{unprivileged}~\cite{10.1145/3524491.3527308}.
In this dataset, each sample consists of the name of the \textit{state} and  the number of crimes associated with each race, i.e., \textit{white}, \textit{black}, \textit{asian}, and \textit{hispanic}. 
For our task, we split races into two groups \textit{White} and \textit{Others} where \textit{Others} groups the crimes of \textit{Black}, \textit{Asian}, and \textit{Hispanic} races
~\cite{DBLP:conf/iclr/BalunovicRV22}. The \textit{privileged} group is the \textit{White} one, and the \textit{unprivileged} is \textit{Others} (i.e., Blacks, Asians, and Hispanics).

Regarding the Adult dataset, we decided to create two settings:\\
(a) - \textit{Adult}: the original dataset where we only discarded the sensitive features \textit{gender} and \textit{marital-status};
(b) - \textit{Adult-debiased}: where we remove all the sensitive features (i.e., gender, age, marital status, and race), and all the features highly correlated with at least one of the sensitive features.\footnote{Both Spearman and Pearson's correlation coefficient greater than 0.35.} As regards the non-sensitive features used for training the models, 
we used: \textit{education num}, \textit{occupation}, \textit{work class}, \textit{capital gain}, \textit{capital loss}, \textit{hours per week}. Furthermore, the feature \textit{work class} has been condensed into three classes: \textit{Private}, \textit{Public}, and \textit{Unemployed}. We replaced the categories in \textit{work class} \textit{Private}, \textit{SelfEmpNotInc}, \textit{SelfEmpInc}, with \textit{Private}, the categories \textit{FederalGov}, \textit{LocalGov}, \textit{StateGov}, with \textit{Private}, and the category \textit{WithoutPay} with \textit{Unemployed}. This diversification ensures coherence with the \textit{fairness under unawareness} setting and makes possible comparisons  with completely biased approaches.
As for the Adult dataset, German contains other sensitive characteristics (i.e., age and race) that we do not include for learning the model to guarantee the \textit{fairness under awareness} setting. Additional information on the datasets, target distribution, sensitive-feature distribution, and ex-ante Statistical Parity are available in Table~\ref{tab:datasetSplit} and Table~\ref{tab:SFdataDisrib}.
\begin{table}
    \centering
    \caption{Overview of relevant dataset information, including sensitive feature distribution, target distribution, name of privileged group, and ex-ante Statistical parity respectively for the Adult, Adult-debiased, German, and Crime datasets.}\label{tab:SFdataDisrib}
    \scriptsize
    \setlength{\tabcolsep}{3.5pt}
\renewcommand{\arraystretch}{1}
\begin{subtable}{\textwidth}
\scriptsize
\begin{tabular}{lllrrr}\hline\toprule
Dataset & $s$ &privileged ($s^+$) &$\Phi(s)^\dagger$ & $\Phi(Y)^{\dagger\dagger}$& ex-ante SP$^*$ \\\hline\midrule
Adult &\textit{gender} &\textit{male} &0.68/0.32 & 0.25/0.75 &0.199 \\
 &\textit{maritalStatus} &\textit{married} &0.48/0.52 &0.25/0.75 &0.378 \\\midrule
Adult-deb. &\textit{gender} &\textit{male} &0.68/0.32 & 0.25/0.75 &0.199 \\
 &\textit{maritalStatus} &\textit{married} &0.48/0.52 &0.25/0.75 & 0.378\\\midrule
 Crime &\textit{race} &\textit{white} &0.58/0.42 & 0.50/0.50 &0.554 \\\midrule
German &\textit{gender} &\textit{male} &0.69/0.31 & 0.70/0.30 &0.075 \\
 &\textit{age} & $>25$ \textit{year} &0.81/0.19 & 0.70/0.30 &0.149 \\

\hline\bottomrule\hline 
\end{tabular}
\end{subtable}


\begin{subtable}{\textwidth}
\scriptsize
\begin{tabular}{lcccc}
\multicolumn{5}{l}{\footnotesize $^\dagger$ Probability distribution of the \textit{privileged} and \textit{unprivileged} group:}  \\
\multicolumn{5}{c}{\footnotesize $\mathds{P}(s^+)/\mathds{P}(s^-)$} \\ 
\multicolumn{5}{l}{\footnotesize $^{\dagger\dagger}$ Probability distribution of the target variable:}\\
\multicolumn{5}{c}{\footnotesize $\mathds{P}(y=1)/\mathds{P}(y=0)$} \\
\multicolumn{5}{l}{\footnotesize $^*$ A priori Statistical Parity, based on Independence criteria:} \\
\multicolumn{5}{c}{\footnotesize $\mathds{P}(y=1 \mid s^+) - \mathds{P}(y=1 \mid s^-)$}
\end{tabular}
\end{subtable}
\vspace{-3em}
\end{table}

\paragraph{Decision Maker.}
To keep the approach as general as possible, we have chosen seven largely adopted learning models to handle the classification task and implement the \textit{Decision Maker}.
In detail, we used: Logistic Regression (LR), Decision Tree (DT), Support-Vector Machines (SVM), LightGBM (LGBM), XGBoost (XGB), Random Forest (RF), and Multi-Layer Perceptron (MLP). We completed our analysis with three \textit{in-processing} debiasing algorithms, Linear Fair Empirical Risk Minimization (LFERM)~\cite{DBLP:conf/nips/DoniniOBSP18}, Adversarial Debiasing (Adv)~\cite{10.1145/3278721.3278779}, and Fair Classification (FairC)~\cite{pmlr-v54-zafar17a}. Those debiasing models have not been casually chosen, but they are consistent with the \textit{fairness under unawareness} setting. Indeed, those models 
can not use sensitive features in the inference phase and 
can be trained only on non-sensitive features (i.e., $\mathbf{x}$), using the sensitive features (i.e., $s_i$) as debiasing constraints.

\paragraph{Counterfactual Generator.}
For the sake of reproducibility and reliability, the counterfactuals are generated by a third-party counterfactual framework.
We used DiCE~\cite{mothilal2020dice,10.1145/3351095.3372850}, an open-source framework developed by Microsoft. DiCE not only offers several strategies for generating counterfactual samples but also is a model-agnostic approach. Furthermore, it is built upon \textit{Proximity}, \textit{Sparsity}, \textit{Diversity}, and \textit{Feasibility} constraints. 
We used the Genetic and KDtree strategies. The number of counterfactuals generated for each negatively predicted sample is equal to 100 (i.e., $|\mathcal{C}_\mathbf{x}|=100$).
We tried to use another agnostic generator, MACE \cite{DBLP:conf/aistats/KarimiBBV20}, but the only compatible models are LR, DT, RF, and MLP (only for a 10-neuron single layer). Moreover, MACE never managed to generate the required number of counterfactuals making the analysis impractical. 
\paragraph{Sensitive-Feature Classifier.}
This component plays a crucial role in our methodology since it allows the system to discover discriminatory models.
For each sensitive feature, a classifier is thus learned. 
We exploited RF, MLP, and XGB for implementing this component.

\paragraph{Metrics.} We evaluate the classifiers' performance with confusion matrix-based Accuracy (ACC) and F1 metrics, the Area~Under~the~Receiver~Operative~Curve~(AUC), and 
separation fairness metric Difference in Equal Opportunity\footnote{\scriptsize $\mathrm{DEO} = |\mathds{P}(\hat{Y}=1 \mid S=1,Y=1) - \mathds{P}(\hat{Y}=1 \mid S=0,Y=1)|$} (DEO).

\paragraph{Split and Hyperparameter Tuning.} The datasets have been divided with the 90/10 train-test split hold-out method. 
We restricted the test set to 10\% due to computational cost of generating counterfactual samples for each data point.
The \textit{Decision Maker} and the \textit{Sensitive-Feature Classifier} models have been tuned on the training set with a Grid Search 5-fold cross-validation methodology, the first optimizing AUC metric and the latter optimizing F1 score to prevent unbalanced predictions on the sensitive feature.

\section{Experiments and discussion}
\label{sec:results}

The Section introduces the experiments and answers the RQs. The analysis of the results will be limited to XGB as $f_s(\cdot)$ due to space constraints. Extended results in Appendix~\ref{sec:AppendixExtendedExperiments}.
\begin{table}\centering
\caption{AUC, Accuracy, and F1 score on the Adult, Adult-debiased, Crime, and German test set of the Sensitive Feature Classifiers. We mark the best-performing method for each metric in bold font.}\label{tab:SFclass}
\scriptsize
\setlength{\tabcolsep}{8pt}
\renewcommand{\arraystretch}{0.9}
\begin{tabular}{lllrrrr}\hline\toprule
& & &  \multicolumn{3}{c}{\textbf{$f_s(\cdot)$}} \\
\cmidrule(l){4-6}
Dataset & $s$ & metric&\textbf{RF} &\textbf{MLP} &\textbf{XGB} \\ \hline \midrule
\multirow{3}{*}{Adult} & &AUC &0.9402 &0.9363 &\textbf{0.9413} \\
&\textit{gender} &ACC &0.8539 &\textbf{0.8559} &0.8463 \\
& &F1 &0.8900 &\textbf{0.8914} &0.8769 \\\midrule
\multirow{3}{*}{Adult-debiased} & &AUC &\textbf{0.8028} &0.8010 &0.7896 \\
&\textit{gender} &ACC &\textbf{0.7482} &0.7480 &0.7444 \\
& &F1 &\textbf{0.8274} &0.8227 &0.8106 \\\midrule
\multirow{3}{*}{Crime} & &AUC &0.9885 &0.9893 &\textbf{0.9910} \\
&\textit{race} &ACC &\textbf{0.9500} &0.9450 &0.9450 \\
& &F1 &\textbf{0.9576} &0.9532 &0.9532 \\\midrule
\multirow{3}{*}{German} & &AUC &0.7106 &0.5091 &\textbf{0.7139} \\
&\textit{gender} &ACC &\textbf{0.7300} &0.6900 &0.6900 \\
& &F1 &\textbf{0.8344} &0.8166 &0.7704 \\
\hline\bottomrule\hline
\end{tabular}
\end{table}


\paragraph{Is there a method for determining whether a dataset contains proxy features or not? (RQ1)}\label{sec:RQ1} This experiment aims to assess how well the sensitive-feature classifier can identify if a subject belongs to the \textit{privileged} or \textit{unprivileged} group, without exploiting sensitive features in the training phase. We trained a  sensitive-feature classifier for each dataset. The investigated sensitive features are \textit{gender} or \textit{race} and results are shown in Table~\ref{tab:SFclass}.
\begin{itemize}[leftmargin=*,topsep=0pt]
\item  
The first observation is that every \textit{Sensitive-Feature Classifier} shows to be accurate for all the datasets. Among them, XGB 
exhibits the best performance in terms of AUC while RF shows the most promising ACC and F1 performance. For the Adult, Adult-debiased, and Crime datasets, MLP performance is comparable to XGB and RF, while shows a low AUC on the German dataset. We recall we seek models with high F1 and AUC since it indicates that the classifiers provide accurate and balanced~predictions.
\item The careful reader may have noticed that, on the Adult-debiased, the sensitive-feature classifiers exhibit the worst performance (in comparison with the original Adult dataset). 
This behavior is due to the debiasing process, as the Adult-debiased dataset has been deprived of features highly correlated with the sensitive ones.  Noteworthy, the prediction capability remains high despite the absence of sensitive and sensitive-correlated features.
\end{itemize}

\noindent \underline{Final comments.}
\textit{Results show that, due to proxy features, it is possible to train a classifier able to predict sensitive characteristics. Moreover, it is still possible to predict sensitive information even when only low correlated features with the sensitive information are available (i.e., Adult-debiased)}.


\paragraph{Does the Fairness Under Unawareness setting ensure that decision biases are avoided? (RQ2)} The \textit{Fairness Under Unawareness} setting aims to ensure fair treatment 
by removing sensitive features from training data. However, as demonstrated previously, it is possible to predict sensitive information due to proxy features. Before evaluating fairness metrics, we evaluated the accuracy performance of the classifiers exploited to implement the \textit{Decision Maker}.
\begin{itemize}[leftmargin=*,topsep=0pt]
    \item Table~\ref{tab:totalTable} indicates that all classifiers work well in terms of accuracy metrics. 
However, as expected, adopting \textit{Fairness Under Unawareness} -- i.e., removing all the sensitive information and, for Adult-debiased, also removing highly correlated features -- has caused a worsening of the performance for all the classifiers (see the comparison between Adult and Adult-debiased results). 
This observation suggests that sensitive and sensitive-correlated information may be ``necessary" to predict the target label correctly. 
Table~\ref{tab:fairnessMetris} reports the fairness evaluation computing the Difference in Equal Opportunity (DEO). 
It is worth noticing that removing the considered sensitive information (i.e., gender and race) has not improved model equity. 
This result shows that not removing proxy features makes the \textit{Fairness Under Unawareness} setting useless since the model can implicitly learn them. 
A clear example is the Adult-debiased dataset, where DEO values are generally better than on the Adult dataset. However, some degree of discrimination is still present due to non-linear proxy features. Furthermore, despite adopting the \textit{in-processing} debiasing constrained optimization, the debiased \textit{Decision Maker} does not seem to improve fairness performance consistently.
\end{itemize}
\underline{Final comments.} 
\textit{The classifiers seem to be affected by discrimination even when the sensitive information is omitted. 
Accordingly, imposing \textit{Fairness Under Unawareness} setting is not sufficient to avoid decision biases and discrimination.}
\input{groupplot/barchart}
\begin{table*}[!htp]\centering
\caption{Accuracy, DEO, and $\Delta$CFlips (\%) metrics with XGB as $f_s(\cdot)$ on the Adult, Adult-debiased, Crime, and German test set. We mark the best-performing method for each metric in bold font.}\label{tab:totalTable}\label{tab:fairnessMetris}
\scriptsize
\setlength{\tabcolsep}{4.5pt}
\renewcommand{\arraystretch}{0.9}
\begin{tabular}{lrrrr|rrrr|rrrr|rrrr}\hline\toprule
&\multicolumn{4}{c}{Adult (\textit{gender})} &\multicolumn{4}{c}{Adult-debiased (\textit{gender})} &\multicolumn{4}{c}{Crime (\textit{race})} &\multicolumn{4}{c}{German (\textit{gender})} \\ \cmidrule(lr){2-5} \cmidrule(lr){6-9} \cmidrule(lr){10-13} \cmidrule(lr){14-17}
& & &\multicolumn{2}{c}{$\Delta\mathrm{CFlips}\downarrow$} & & &\multicolumn{2}{c}{$\Delta\mathrm{CFlips}\downarrow$} & & &\multicolumn{2}{c}{$\Delta\mathrm{CFlips}\downarrow$} & & &\multicolumn{2}{c}{$\Delta\mathrm{CFlips}\downarrow$} \\ \cmidrule(lr){4-5}\cmidrule(lr){8-9}\cmidrule(lr){12-13}\cmidrule(lr){16-17}
$f(\cdot)$ &ACC$\uparrow$ &DEO$\downarrow$ &Genetic &\multicolumn{1}{c}{KDtree} &ACC$\uparrow$ &DEO$\downarrow$ &Genetic &\multicolumn{1}{c}{KDtree} &ACC$\uparrow$ &DEO$\downarrow$ &Genetic &\multicolumn{1}{c}{KDtree} &ACC$\uparrow$ &DEO$\downarrow$ &Genetic &\multicolumn{1}{c}{KDtree} \\ \hline
\midrule
LR &0.8099 &0.0546 &67.05 &76.03 &0.7367 &0.0695 &40.02 &\textbf{8.37} &\textbf{0.8700} &0.3294 &81.70 &78.76 &0.7600 &0.1400 &11.89 &35.65 \\
DT &0.8161 &0.0760 &70.23 &77.14 &0.8017 &0.0492 &39.65 &12.74 &0.8200 &0.4039 &72.80 &66.25 &0.7600 &0.0500 &21.61 &\textbf{22.43} \\
SVM &0.8541 &0.0644 &73.99 &79.35 &0.8061 &0.0353 &\textbf{18.27} &11.66 &\textbf{0.8700} &0.3843 &79.17 &75.15 &0.7600 &0.0300 &38.64 &37.29 \\
LGBM &0.8658 &0.0379 &70.87 &78.40 &0.8371 &0.0470 &56.16 &59.11 &0.8400 &0.2824 &78.22 &76.52 &0.7500 &0.1900 &\textbf{3.54} &28.54 \\
XGB &\textbf{0.8698} &0.0635 &70.40 &78.42 &\textbf{0.8375} &0.0400 &69.02 &59.22 &0.8500 &0.2941 &78.89 &76.79 &\textbf{0.7900} &0.0400 &7.85 &32.37 \\
RF &0.8534 &0.0216 &73.09 &76.68 &0.8267 &0.0703 &71.12 &83.87 &0.8400 &0.2824 &70.23 &68.19 &0.7600 &0.0800 &15.44 &37.17 \\
MLP &0.8494 &0.0529 &71.32 &77.95 &0.8156 &\textbf{0.0173} &92.38 &89.37 &0.8650 &0.3294 &76.41 &75.22 &0.7600 &0.0300 &29.07 &29.00 \\
LFERM &0.8428 &\textbf{0.0194} &32.40 &68.25 &0.7953 &0.0179 &39.67 &60.55 &0.8400 &0.2941 &68.68 &63.69 &0.7200 &\textbf{0.0200} &24.24 &23.23 \\
ADV &0.8512 &0.1399 &\textbf{7.96} &53.90 &0.8196 &0.0326 &33.17 &85.81 &0.8100 &0.1882 &74.09 &74.12 &0.7300 &0.2200 &18.46 &31.51 \\
FairC &0.8395 &0.2451 &38.72 &\textbf{36.27} &0.8054 &0.0529 &80.12 &72.24 &0.7500 &\textbf{0.1373} &\textbf{27.06} &\textbf{18.40} &0.7400 &0.0500 &23.94 &30.81 \\
\hline
\bottomrule
\hline
\end{tabular}
\end{table*}

\paragraph{Is counterfactual reasoning effective to discover decision biases? (RQ3)}

\noindent This experiment aims to unveil potential decision biases by counterfactual reasoning. Figure~\ref{fig:totalFigures} reports the $\mathrm{CFlips}$ values (see Definition~\ref{eq:CFlips}) for each classifier and category -- i.e., \textit{privileged}, see Eq.~\ref{eq:CFlipsPriv}, and \textit{unprivileged}, see Eq.~\ref{eq:CFlipsUnpriv}, with XGB as $f_s(\cdot)$. The metric tells us how frequently a change in the decision (from negative to positive) for a sample is followed by a change in the sensitive-feature classification (e.g., from \textit{female} to \textit{male} and~vice-versa).
\begin{itemize}[leftmargin=*,topsep=0pt]
    \item The proposed metric seems to operate as expected since some hidden discriminatory behaviors emerge. For instance, the counterfactuals belonging to the \textit{unprivileged} category, i.e., \textit{female} or \textit{others}, have a much higher $\mathrm{CFlips}$ than counterfactuals of \textit{privileged} samples, i.e., \textit{male} or \textit{white}. 
This high percentage of flips for the unprivileged category means that the counterfactuals for the female (and "others" race) group must show male (and white) characteristics to get a positive decision.
In this respect, Adult and Crime are characterized by the highest CFlips values and the largest difference between the \textit{privileged} and \textit{unprivileged} groups (see Table~\ref{tab:fairnessMetris}).
We underline that $\mathrm{CFlips}$ and $\Delta\mathrm{CFlips}$ results complement, they explain (e.g., highlighting if privileged group characteristics lead to a positive outcome) but also overturn the DEO metric in Table~\ref{tab:fairnessMetris}, shedding light on how the discriminatory classifiers work.

\item The debiasing models perform particularly well (see the DEO metric in Table~\ref{tab:fairnessMetris}) on datasets where sensitive features can be easily identified, i.e., the datasets characterized by accurate sensitive-feature classifiers. Notwithstanding, the $\Delta\mathrm{CFlips}$ in Table~\ref{tab:fairnessMetris} highlights that a certain degree of discrimination persists for the datasets i) with features with a low correlation with the sensitive features or ii) composed by just a few samples.
For instance, for the Crime dataset, even though all the sensitive-feature classifiers exhibit high accuracy, only FairC succeeds in decreasing discrimination at the expense of a significant loss in accuracy.

\item 
The Adult-debiased dataset shows a smaller number of CFlips than Adult, especially for LR and SVM. However, even the most accurate XGB and LGBM show an evident discrepancy between the \textit{privileged} and \textit{unprivileged} groups. This might indicate that both classifiers learned correlations between the proxy features and the target.
The German dataset has a similar trend with MLP and SVM as the most affected by unfairness.
Finally, German's small test set size and the low accuracy of XGB as $f_s(\cdot)$ drive MLP, and LFERM to have 0 flips in the \textit{female} category.

\begin{figure}[t]
\pgfplotsset{tiny,width=3.85cm,compat=1.17, title style={font=\tiny}}
\begin{subfigure}{1\textwidth}
\tiny
\begin{tikzpicture}
	\begin{axis}[
	legend columns = -1,
	legend entries = {LR, DT, SVM, LGBM, XGB, RF, MLP, LFERM, ADV, FairC},
	legend to name = name,
    title={Adult \textit{gender}},
    xlabel={$\mid \mathcal{C}_\mathbf{x} \mid$},
    yticklabel={\pgfmathprintnumber\tick\%},
    ylabel={$\Delta$CFlips($\mathcal{X}^{-}$)},
    every axis/.append style={font=\tiny},
    ymin=0, ymax=100,
]
\addplot[color=clrLR,
    mark=none,] table [x=i, y=j, col sep=comma ] {csv/AdultG/LR_D.csv};
\addplot[color=clrDT,
    mark=none,] table [x=i, y=j, col sep=comma ] {csv/AdultG/DT_D.csv};
\addplot[color=clrSVM,
    mark=none,] table [x=i, y=j, col sep=comma ] {csv/AdultG/SVM_D.csv};
\addplot[color=clrLGB,
    mark=none,] table [x=i, y=j, col sep=comma ] {csv/AdultG/LGB_D.csv};
\addplot[color=clrXGB,
    mark=none,] table [x=i, y=j, col sep=comma ] {csv/AdultG/XGB_D.csv};
\addplot[color=clRF,
    mark=none,] table [x=i, y=j, col sep=comma ] {csv/AdultG/RF_D.csv};
\addplot[color=clMLP,
    mark=none,] table [x=i, y=j, col sep=comma ] {csv/AdultG/MLP_D.csv};
\addplot[color=clLFERM,
    mark=none,] table [x=i, y=j, col sep=comma ] {csv/AdultG/lferm_D.csv};
\addplot[color=clADV,
    mark=none,] table [x=i, y=j, col sep=comma ] {csv/AdultG/Adv_D.csv};
\addplot[color=clZafar,
    mark=none,] table [x=i, y=j, col sep=comma ] {csv/AdultG/zafar_D.csv};
    \end{axis}
\end{tikzpicture}
\begin{tikzpicture}
	\begin{axis}[
    title={Adult \textit{maritalStatus}},
    yticklabel={\empty},
    xlabel={$\mid \mathcal{C}_\mathbf{x} \mid$},
    ymin=0, ymax=100,
]
\addplot[color=clrLR,
    mark=none,] table [x=i, y=j, col sep=comma ] {csv/AdultM/LR_D.csv};
\addplot[color=clrDT,
    mark=none,] table [x=i, y=j, col sep=comma ] {csv/AdultM/DT_D.csv};
\addplot[color=clrSVM,
    mark=none,] table [x=i, y=j, col sep=comma ] {csv/AdultM/SVM_D.csv};
\addplot[color=clrLGB,
    mark=none,] table [x=i, y=j, col sep=comma ] {csv/AdultM/LGB_D.csv};
\addplot[color=clrXGB,
    mark=none,] table [x=i, y=j, col sep=comma ] {csv/AdultM/XGB_D.csv};
\addplot[color=clRF,
    mark=none,] table [x=i, y=j, col sep=comma ] {csv/AdultM/RF_D.csv};
\addplot[color=clMLP,
    mark=none,] table [x=i, y=j, col sep=comma ] {csv/AdultM/MLP_D.csv};
\addplot[color=clLFERM,
    mark=none,] table [x=i, y=j, col sep=comma ] {csv/AdultM/lferm_D.csv};
\addplot[color=clADV,
    mark=none,] table [x=i, y=j, col sep=comma ] {csv/AdultM/Adv_D.csv};
\addplot[color=clZafar,
    mark=none,] table [x=i, y=j, col sep=comma ] {csv/AdultM/zafar_D.csv};
    \end{axis}
\end{tikzpicture}
\begin{tikzpicture}
	\begin{axis}[
    title={Crime \textit{race}},
    xlabel={$\mid \mathcal{C}_\mathbf{x} \mid$},
    yticklabel={\empty},
    ymin=0, ymax=100,
]
\addplot[color=clrLR,
    mark=none,] table [x=i, y=j, col sep=comma ] {csv/Crime/LR_D.csv};
\addplot[color=clrDT,
    mark=none,] table [x=i, y=j, col sep=comma ] {csv/Crime/DT_D.csv};
\addplot[color=clrSVM,
    mark=none,] table [x=i, y=j, col sep=comma ] {csv/Crime/SVM_D.csv};
\addplot[color=clrLGB,
    mark=none,] table [x=i, y=j, col sep=comma ] {csv/Crime/LGB_D.csv};
\addplot[color=clrXGB,
    mark=none,] table [x=i, y=j, col sep=comma ] {csv/Crime/XGB_D.csv};
\addplot[color=clRF,
    mark=none,] table [x=i, y=j, col sep=comma ] {csv/Crime/RF_D.csv};
\addplot[color=clMLP,
    mark=none,] table [x=i, y=j, col sep=comma ] {csv/Crime/MLP_D.csv};
\addplot[color=clLFERM,
    mark=none,] table [x=i, y=j, col sep=comma ] {csv/Crime/lferm_D.csv};
\addplot[color=clADV,
    mark=none,] table [x=i, y=j, col sep=comma ] {csv/Crime/Adv_D.csv};
\addplot[color=clZafar,
    mark=none,] table [x=i, y=j, col sep=comma ] {csv/Crime/zafar_D.csv};
    \end{axis}
\end{tikzpicture}
\end{subfigure}
\pgfplotsset{tiny,width=2cm,compat=1.17}
\begin{subfigure}{1\textwidth} 

\begin{adjustbox}{width=0.48\textwidth}
\ref{name}
\end{adjustbox}
\end{subfigure}
\caption{Ablation study at different number of generated CF (i.e. $\mid \mathcal{C}_\mathbf{x} \mid$) for each sample with Genetic strategy. For Adult, we also investigate a second sensitive feature (i.e., \textit{marital-status}).
}
    \label{fig:ablationAdult}
\vspace{-1em}
\end{figure}
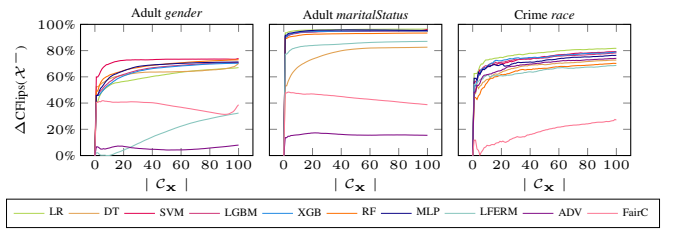

\item Figure~\ref{fig:ablationAdult} analyzes the impact of the number of generated counterfactuals and the validity of the metric $\Delta\mathrm{CFlips}$ when different sensitive features are present in the same dataset. The figure shows that the values of $\Delta\mathrm{CFlips}$ are stable and reliable if the metric is computed on at least 20 counterfactuals for each sample. 
A different behavior can be observed for LFERM in Adult-gender and FairC in Crime-race that start with an optimal performance, and then their $\Delta\mathrm{CFlips}$ increases linearly. This trend needs further investigation, but it could be related to the higher number of counterfactuals.
Indeed, with several counterfactuals, some could be farther from the original sample, and this distance probably entails a higher $\Delta\mathrm{CFlips}$. However, in fairness research, measuring the distance of a sample from the decision boundary of the classifier is a timely challenge~\cite{DBLP:conf/innovations/DworkHPRZ12}. It is worth mentioning that the analysis can be conducted for different features on the same dataset, even when it contains more than one sensitive feature (see \textit{gender} and \textit{maritalStatus} plots for Adult dataset).

\begin{table}\centering
\caption{Leveraging different \textit{sensitive-feature classifiers} (i.e., RF, MLP, and XGB) and testing CFlips metrics. As expected, for Adult and Crime datasets whose classifiers perform accurately, the differences in the predictions of CFlips are negligible.}\label{tab:differenceCFlips}
\scriptsize
\setlength{\tabcolsep}{2.5pt}
\renewcommand{\arraystretch}{0.8}
\begin{tabular}{lllrr|rr|rr}\hline\toprule
& & &\multicolumn{6}{c}{$\Delta\mathrm{CFlips}$} \\ \cmidrule(lr){4-9}
& & &\multicolumn{2}{c}{\textbf{RF}} &\multicolumn{2}{c}{\textbf{MLP}} &\multicolumn{2}{c}{\textbf{XGB}} \\  \cmidrule(lr){4-5} \cmidrule(lr){6-7} \cmidrule(lr){8-9} 
Dataset &$s$ &$f(\cdot)$ &Genetic &\multicolumn{1}{c}{KDtree} &Genetic &\multicolumn{1}{c}{KDtree} &Genetic &\multicolumn{1}{c}{KDtree} \\ \hline \midrule
\multirow{10}{*}{Adult} &\multirow{10}{*}{\textit{gender}} &LR &67.35 &76.09 &66.75 &75.28 &67.05 &76.03 \\
& &DT &70.10 &77.16 &70.07 &77.14 &70.23 &77.14 \\
& &SVM &74.03 &79.33 &73.99 &79.82 &73.99 &79.35 \\
& &LGBM &70.72 &78.40 &70.26 &77.57 &70.87 &78.40 \\
& &XGB &70.34 &78.59 &69.54 &77.42 &70.40 &78.42 \\
& &RF &73.24 &77.01 &72.25 &75.68 &73.09 &76.68 \\
& &MLP &71.17 &77.86 &71.24 &77.80 &71.32 &77.95 \\
& &LFERM &32.15 &68.13 &32.33 &67.86 &32.40 &68.25 \\
& &ADV &\textbf{4.36} &53.70 &\textbf{4.25} &53.80 &\textbf{7.96} &53.90 \\
& &FairC &43.09 &\textbf{35.86} &43.51 &\textbf{35.78} &38.72 &\textbf{36.27} \\ \midrule
\multirow{10}{*}{Crime} &\multirow{10}{*}{\textit{race}} &LR &80.56 &78.33 &82.57 &86.31 &81.03 &78.76 \\
& &DT &72.48 &66.04 &74.13 &71.76 &72.59 &66.25 \\
& &SVM &79.02 &75.31 &83.60 &83.36 &78.29 &75.15 \\
& &LGB &78.23 &76.65 &82.07 &85.11 &77.82 &76.52 \\
& &XGB &78.93 &76.84 &81.46 &85.35 &78.52 &76.79 \\
& &RF &70.27 &68.12 &72.87 &73.84 &69.32 &68.19 \\
& &MLP &76.90 &74.66 &82.32 &82.31 &76.04 &75.22 \\
& &LFERM &68.42 &63.85 &71.16 &71.44 &68.32 &63.69 \\
& &ADV &74.51 &74.10 &74.42 &81.80 &73.62 &74.12 \\
& &FairC &\textbf{25.50} &\textbf{17.60} &\textbf{31.77} &\textbf{25.13} &\textbf{25.46} &\textbf{18.40} \\
\hline \bottomrule \hline
\end{tabular}
\end{table}
\item Table~\ref{tab:differenceCFlips} extends the analysis to different $f_s(\cdot)$ (for the sake of space, the previous experiments adopted XGB as $f_s(\cdot)$). The results show that, also for Adult and Crime, the proposed metric is stable and reliable.

\item The counterfactual generation strategies reveal some important findings: there exist similar real samples (i.e., similar people) for which the switch to the privileged group led to a positive outcome. Indeed, KDtree searches among dataset samples -- i.e., $\mathbf{c}_{\mathbf{x}} \in \mathcal{D}$ -- thus $\mathrm{CFlips} \ge 0$ is critical. Genetic strategy, instead, analyzes the unexplored space and confirms the discriminatory behavior.
\end{itemize}
\underline{Final comments.} \textit{In the various plots emerges that the \textit{unprivileged} samples, to achieve favorable decisions, must take on the characteristics of \textit{privileged}~samples. The results demonstrate that counterfactual reasoning effectively discovers decision biases and complements SOTA fairness metrics.}

\begin{figure}
\vspace{-1em}
\tiny
\centering
\pgfplotsset{tiny,width=3.5cm,compat=1.17, title style={font=\tiny}}
\begin{tikzpicture}
\begin{axis}[ 
xbar, xmin=-1.1,xmax=1.1,
xlabel={\scriptsize $\rho(\epsilon,\delta)$ },
symbolic y coords={%
    {Priv-house-serv (Occupation)},
    {Craft-repair (Occupation)},
    {Exec-managerial (Occupation)},
    {Adm-Clerical (Occupation)},
    {education-num},
    {hours per week}},
    bar width=0.25cm,
    height = 3.5cm,
    width = 6cm,
    xtick={-1,-0.5,0,0.5,1},
ytick=data,
nodes near coords, 
nodes near coords align={horizontal},
ytick=data,
]
\addplot coordinates {
    (-0.287845,{Priv-house-serv (Occupation)}) 
    (0.309775,{Craft-repair (Occupation)}) 
    (0.346542,{Exec-managerial  (Occupation)})
    (-0.521061,{Adm-Clerical (Occupation)})
    (0.541314,{education-num}) 
    (0.566165,{hours per week})};
\end{axis}
\end{tikzpicture}
\caption{Top-6 most correlated features with a \textit{gender} Flip (i.e., $\rho(\epsilon,\delta)$) on the Adult-debiased dataset with Genetic strategy as $g(\mathbf{x})$ and MLP as both $f(\cdot)$ and $f_s(\cdot)$.
}
    \label{fig:explainCorr}
    \vspace{-3.5em}
\end{figure}
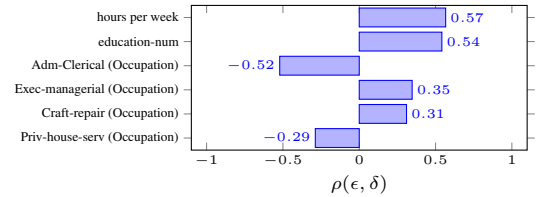

\paragraph{Is it possible to define a strategy for identifying the proxy features? (RQ4)}
In RQ1, we highlighted how it is possible to determine if a dataset contains proxy features. Here, we define a strategy to identify them in the dataset.
A counterfactual $\mathbf{c}_\mathbf{x}$ can be seen as a perturbation from a starting sample $\mathbf{x}$ of a quantity $\mathbf{\epsilon}$ (i.e., $\mathbf{c}_\mathbf{x} = \mathbf{x}+\mathbf{\epsilon}$). For a numerical or ordinal feature $i$, $\epsilon_i$ can be expressed as the difference between the counterfactual and the feature of the sample  $c_{x_i} - x_i$. For a categorical feature $j$, $\epsilon_j$ can be expressed in a \textit{one-hot encoding} form as -1 to the category that is removed and 1 to the category that is engaged. Let be $\delta$ the difference between the posterior conditional probability of predicting a counterfactual sample and 
the original sample as belonging to the privileged group (i.e., $\delta = \mathds{P}(f_s(\mathbf{c}_{\mathbf{x}})=1|\mathbf{c}_\mathbf{x}) - \mathds{P}(f_s(\mathbf{x})=1|\mathbf{x})$). 
We can identify the most influential features for $f_s(\cdot)$ evaluating the Pearson correlation between $\epsilon$ and $\delta$: $\rho(\epsilon, \delta)$ (a demonstrative example in Table~\ref{tab:Correlation}).
In Figure~\ref{fig:explainCorr} we can find the top-6 most correlated features with a Flip in $f_s(\cdot)$ with MLP as $f(\cdot)$ decision boundary for the generation of $\mathbf{c}_\mathbf{x}$ and also as $f_s(\cdot)$ for the Adult-debiased dataset. A negatively correlated feature (e.g., \textit{Adm-Clerical}) is a feature that has an opposite direction respect $\mathds{P}(\hat{s}=1 \mid \mathbf{x})$ while a positively correlated one (e.g., \textit{hours per weeek}) has the same direction. 
\begin{table}\centering
\caption{Demonstrative example of $\rho$ computation based on $\epsilon$ and $\delta$ for a numeric and categorical feature of Adult-debiased dataset. }\label{tab:Correlation}
\scriptsize
\setlength{\tabcolsep}{1pt}
\renewcommand{\arraystretch}{0.2}
\begin{tabular}{lrrrrr|lr}\hline\toprule
&numeric &ordinal&\multicolumn{3}{c}{Category (\textit{workclass})} & & \textit{gender} \\\cmidrule(l){2-2}\cmidrule(l){3-3}\cmidrule(lr){4-6}\cmidrule(l){8-8}
&\textit{capital gain} & \textit{education-n.}&\textit{Private} &\textit{Public} &\multicolumn{1}{c}{\textit{Unemployed}} & &$\mathds{P}(\hat{s}=1|x)$ \\\hline\midrule
$\mathbf{c}_{\mathbf{x}_1}$ &5000 & 6&1 &0 &0 &$f_s(\mathbf{c}_{\mathbf{x}_1})$&0.7 \\
$\mathbf{x}_1$ &2000 & 2&0 &0 &1 &$f_s(\mathbf{x}_1)$ &0.1 \\ \midrule
$\mathbf{\epsilon_{\mathbf{x}_1}}$ &3000 &4 &1 &0 &-1 &$\delta_{\mathbf{x}_1}$&0.6 \\ \midrule $\mathbf{c}_{\mathbf{x}_2}$ &2800 & 5&0 &1 &0 &$f_s(\mathbf{c}_{\mathbf{x}_2})$&0.3 \\
$\mathbf{x}_2$ &600 &5&1 &0 &0 &$f_s(\mathbf{x}_2)$ &0.7 \\ \midrule
$\mathbf{\epsilon_{\mathbf{x}_2}}$ &2200 & 0 &-1 &1 &0 &$\delta_{\mathbf{x}_2}$&-0.4 \\ \midrule
\multicolumn{7}{c}{$\dots$}\\ \midrule
$\mathbf{\epsilon_{\mathbf{x}_3}}$ &1200 &-1&0 &1 &-1 &$\delta_{\mathbf{x}_3}$&-0.6 \\ \midrule
$\rho(\epsilon,\delta)$ & 0.91&0.93 &0.78& -0.99& \multicolumn{1}{r}{-0.36} &  \\ 
\hline\bottomrule\hline
\end{tabular}
\end{table}

\section{Conclusion and Future Work}
\label{sec:conclusion}
In the context of \textit{fairness under unawareness}, where the sensitive information is omitted, \textit{Decision Makers} can often be biased as they leverage proxy features of sensitive ones.
In this work, we present a novel methodology to detect bias in \textit{Decision-Making} models by analyzing 
the sensitive behavior of (counterfactual) samples belonging to the positive side of the decision boundary.
In detail, we exploit three different counterfactual generation strategies to do so. Specifically, the newly generated counterfactuals could belong to another sensitive group, thus suggesting potential discrimination in the decision process. 
To comprehensively assess the proposed methodology, we conducted experiments with ten decision makers (including state-of-the-art debiasing models) and three sensitive-feature classifiers. To measure the extent of the discriminatory behavior, we introduce a new metric to find how many counterfactuals belong to another sensitive group. 
The contribution of this work is manifold: (i) we demonstrate that \textit{fairness under unawareness} assumption is not sufficient to mitigate bias, (ii) we propose a methodology for the bias auditing task, (iii) we show that counterfactual reasoning is an effective methodology to unveil the bias, and (iv) we define a procedure to identify proxy features leveraging counterfactual reasoning. 
In the future, we plan to define a strategy to generate fair and actionable counterfactual samples with the aim of developing a model that could be effectively fair in the context of \textit{fairness under unawareness}.

\ack
This work was partially supported by the following projects: Secure Safe Apulia,
MISE CUP: I14E20000020001 CTEMT - Casa delle Tecnologie Emergenti Comune di Matera, CT\_FINCONS\_III, OVS Fashion Retail Reloaded, LUTECH DIGITALE 4.0, KOINÈ.

\bibliography{ecai}

\newpage
\clearpage
\newpage

\appendix
\newpage
\section{Appendix}\label{sec:Appendix}
\subsection{EXPERIMENTAL SETUP}\label{sec:AppendixSetup}
This section provides a detailed description of our experimental settings.

\paragraph{Datasets.} Experiments have been conducted on three state-of-the-art (SOTA) datasets, used as benchmarks in several  works~\cite{DBLP:conf/iclr/BalunovicRV22,DBLP:conf/nips/DoniniOBSP18,DBLP:conf/kdd/PedreschiRT08,Das33}. These  are: Adult\cite{kohavi_becker_1996}, a real-world dataset used for income prediction\footnote{\underline{Adult}: \url{https://archive.ics.uci.edu/ml/datasets/adult}},  German\cite{german_hoffman}, a real-world dataset for default prediction\footnote{\underline{German}:~\url{https://archive.ics.uci.edu/ml/datasets/statlog+(german+credit+data)}}, and Crime~\cite{crime_data}, a real-world Census dataset for violent state prediction\footnote{\underline{Crime}:~\url{https://archive.ics.uci.edu/ml/datasets/US+Census+Data+(1990)}} (i.e., a state is violent if the number of crime in a state is higher with respect to the median($| \mathcal{C}_\mathbf{x} |$) of all the states).
For Adult and German, the sensitive attribute we considered is \textit{gender}, with \textit{male} and \textit{female} corresponding to the \textit{privileged} and \textit{unprivileged}~group respectively. For the Crime dataset, the sensitive attribute we considered is the \textit{race} that indicates the race with the largest number of crimes committed in a specific state. As a second sensitive feature, for Adult, we chose \textit{maritalStatus}, with \textit{married} and \textit{not married} as \textit{privileged} and \textit{unprivileged}, and for German, we chose age as $>25$ \textit{years} and $<=25$ \textit{years} as \textit{privileged} and \textit{unprivileged}
In this dataset, each sample consists of the name of the \textit{state} and  the number of crimes associated with each race, i.e., \textit{white}, \textit{black}, \textit{asian}, and \textit{hispanic}. 
For our task, we split races into two groups \textit{White} and \textit{Others} where \textit{Others} groups the crimes of \textit{Black}, \textit{Asian}, and \textit{Hispanic} races. This 
reproduces the setting of \cite{DBLP:conf/iclr/BalunovicRV22}. The \textit{privileged} group is the \textit{White} one, and the \textit{unprivileged} is \textit{Others} (i.e., Blacks, Asians, and Hispanics).

Regarding the Adult dataset, we decided to create two different settings:\\
(a) - \textit{Adult}: the original dataset where we only discarded the sensitive features \textit{gender} and \textit{marital-status};
(b) - \textit{Adult-debiased}: where we remove all the sensitive features (i.e., gender, age, marital status, and race), and all the features highly correlated with at least one of the sensitive features\footnote{Pearson's correlation coefficient greater than 0.35.}. As regards the non-sensitive features used for training the models, 
we used: \textit{education num}, \textit{occupation}, \textit{work class}, \textit{capital gain}, \textit{capital loss}, \textit{hours per week}. Furthermore, the feature \textit{work class} has been condensed into three classes: \textit{Private}, \textit{Public}, and \textit{Unemployed}. We replaced the categories in \textit{work class} \textit{Private}, \textit{SelfEmpNotInc}, \textit{SelfEmpInc}, with \textit{Private}, the categories \textit{FederalGov}, \textit{LocalGov}, \textit{StateGov}, with \textit{Private}, and the category \textit{WithoutPay} with \textit{Unemployed}. This diversification ensures coherence with the \textit{fairness under unawareness} setting and makes possible comparisons  with completely biased approaches.
As for the Adult dataset, German contains other sensitive characteristics (i.e., age and race) that we do not include for learning the model to guarantee the \textit{fairness under awareness} setting. Additional information on the datasets, target distribution, sensitive-feature distribution, and ex-ante Statistical Parity are available in Table~\ref{tab:datasetSplit} and Table~\ref{tab:SFdataDisrib}.

\paragraph{Decision Maker.}
To keep the approach as general as possible, we have chosen four largely adopted learning models to tackle the classification task and implement the \textit{decision-maker}.
In detail, we used: Logistic Regression (LR), Decision Tree (DT), Support-Vector Machines (SVM), LightGBM\footnote{\underline{LGBM}: \url{https://github.com/microsoft/LightGBM}} (LGBM), XGBoost\footnote{\underline{XGB}: \url{https://github.com/dmlc/xgboost}} (XGB), Random Forest (RF), and Multi-Layer Perceptron (MLP)\footnote{\underline{LR}, \underline{DT}, \underline{SVM}, \underline{RF}, \underline{MLP}: \url{https://scikit-learn.org/}}. We extended our analysis with three in-processing debiasing algorithms, Linear Fair Empirical Risk Minimization (LFERM)~\footnote{\underline{LFERM}: \url{https://github.com/jmikko/fair_ERM}}~\cite{DBLP:conf/nips/DoniniOBSP18}, Adversarial Debiasing (Adv)~\footnote{\underline{Adv}: \url{https://github.com/Trusted-AI/AIF360}}~\cite{10.1145/3278721.3278779}, and Fair Classification (FairC)\footnote{\underline{Fairc}: \url{https://github.com/mbilalzafar/fair-classification}}~\cite{pmlr-v54-zafar17a}. Debiasing models are not chosen casually since consistent with the \textit{fairness under unawareness} setting. Indeed, they are models that may not use sensitive features in the inference phase and may only be trained on non-sensitive features (i.e., $\mathbf{x}$) using the sensitive features (i.e., $s_i$) as debiasing constraints. 

\paragraph{Counterfactual Generator.}

\begin{table}\centering
\caption{
Statistics of three-generation strategy on the test set, showing samples with at least one counterfactual (CF)
and the median of CF.
\textbf{E.C.T.}: 
\textit{extensive computation time}, \textbf{N.C.}: \textit{not compatible} parameter.
}\label{tab:StatisticsCF}
\scriptsize
\setlength{\tabcolsep}{3.5pt}
\renewcommand{\arraystretch}{0.8}
\begin{tabular}{llllrrrr}\hline\toprule
& & & &\multicolumn{3}{c}{\textbf{CF generation strategy}} \\\cmidrule(lr){5-7}
Dataset &$s$ &model &Statistics &Genetic &KDtree &MACE \\ \midrule\hline
\multirow{8}{*}{Adult} &\multirow{8}{*}{gender} &\multirow{2}{*}{LR} &$|\mathcal{D}|$ with $|\mathcal{C}_\mathbf{x}|>0$&4523 &4523 &2854 \\
& & &median($| \mathcal{C}_\mathbf{x} |$) &100 &100 &8 \\\cmidrule(lr){3-7}
& &\multirow{2}{*}{DT} &$|\mathcal{D}|$ with $|\mathcal{C}_\mathbf{x}|>0$&4523 &4523 &3857 \\
& & &median($| \mathcal{C}_\mathbf{x} |$) &100 &100 &7 \\\cmidrule(lr){3-7}
& &\multirow{2}{*}{RF} &$|\mathcal{D}|$ with $|\mathcal{C}_\mathbf{x}|>0$&4523 &4523 &\textbf{E.C.T.} \\
& & &median($| \mathcal{C}_\mathbf{x} |$) &100 &100 &\textbf{E.C.T.} \\\cmidrule(lr){3-7}
& &\multirow{2}{*}{MLP} &$|\mathcal{D}|$ with $|\mathcal{C}_\mathbf{x}|>0$&4523 &4523 &\textbf{N.C.} \\
& & &median($| \mathcal{C}_\mathbf{x} |$) &100 &100 &\textbf{N.C.} \\ \midrule
\multirow{8}{*}{Adult-deb.} &\multirow{8}{*}{gender} &\multirow{2}{*}{LR} &$|\mathcal{D}|$ with $|\mathcal{C}_\mathbf{x}|>0$&4523 &4523 &2857 \\
& & &median($| \mathcal{C}_\mathbf{x} |$) &100 &95 &5 \\\cmidrule(lr){3-7}
& &\multirow{2}{*}{DT} &$|\mathcal{D}|$ with $|\mathcal{C}_\mathbf{x}|>0$&4523 &4523 &4183 \\
& & &median($| \mathcal{C}_\mathbf{x} |$) &98 &100 &5 \\\cmidrule(lr){3-7}
& &\multirow{2}{*}{RF} &$|\mathcal{D}|$ with $|\mathcal{C}_\mathbf{x}|>0$&4523 &4523 &3929 \\
& & &median($| \mathcal{C}_\mathbf{x} |$) &98 &81 &5 \\\cmidrule(lr){3-7}
& &\multirow{2}{*}{MLP} &$|\mathcal{D}|$ with $|\mathcal{C}_\mathbf{x}|>0$&4523 &4523 &\textbf{N.C.} \\
& & &median($| \mathcal{C}_\mathbf{x} |$) &98 &77 &\textbf{N.C.} \\\midrule
\multirow{8}{*}{Crime} &\multirow{8}{*}{race} &\multirow{2}{*}{LR} &$|\mathcal{D}|$ with $|\mathcal{C}_\mathbf{x}|>0$&200 &200 &0 \\
& & &median($| \mathcal{C}_\mathbf{x} |$) &100 &100 &0 \\\cmidrule(lr){3-7}
& &\multirow{2}{*}{DT} &$|\mathcal{D}|$ with $|\mathcal{C}_\mathbf{x}|>0$&200 &200 &\textbf{E.C.T.} \\
& & &median($| \mathcal{C}_\mathbf{x} |$) &100 &100 &\textbf{E.C.T.} \\\cmidrule(lr){3-7}
& &\multirow{2}{*}{RF} &$|\mathcal{D}|$ with $|\mathcal{C}_\mathbf{x}|>0$&200 &200 &\textbf{E.C.T.} \\
& & &median($| \mathcal{C}_\mathbf{x} |$) &100 &100 &\textbf{E.C.T.} \\\cmidrule(lr){3-7}
& &\multirow{2}{*}{MLP} &$|\mathcal{D}|$ with $|\mathcal{C}_\mathbf{x}|>0$&200 &200 &\textbf{N.C.} \\
& & &median($| \mathcal{C}_\mathbf{x} |$) &100 &100 &\textbf{N.C.} \\\midrule
\multirow{8}{*}{German} &\multirow{8}{*}{gender} &\multirow{2}{*}{LR} &$|\mathcal{D}|$ with $|\mathcal{C}_\mathbf{x}|>0$&100 &100 &33 \\
& & &median($| \mathcal{C}_\mathbf{x} |$) &100 &100 &7 \\\cmidrule(lr){3-7}
& &\multirow{2}{*}{DT} &$|\mathcal{D}|$ with $|\mathcal{C}_\mathbf{x}|>0$&100 &100 &19 \\
& & &median($| \mathcal{C}_\mathbf{x} |$) &100 &100 &9 \\\cmidrule(lr){3-7}
& &\multirow{2}{*}{RF} &$|\mathcal{D}|$ with $|\mathcal{C}_\mathbf{x}|>0$&100 &100 &16 \\
& & &median($| \mathcal{C}_\mathbf{x} |$) &98 &100 &9 \\\cmidrule(lr){3-7}
& &\multirow{2}{*}{MLP} &$|\mathcal{D}|$ with $|\mathcal{C}_\mathbf{x}|>0$&100 &100 &\textbf{N.C.} \\
& & &median($| \mathcal{C}_\mathbf{x} |$) &99 &100 &\textbf{N.C.} \\
\hline\bottomrule\hline
\end{tabular}\vspace{-1.7em}
\end{table}
\begin{table}\centering
\caption{Complete results on the Adult, Adult-debiased, Crime, and German test set of the Sensitive Feature Classifiers.}\label{tab:extSFclf}
\scriptsize
    \setlength{\tabcolsep}{3.5pt}
\renewcommand{\arraystretch}{0.8}
\begin{tabular}{lllrrrr}\hline\toprule
& & &  \multicolumn{3}{c}{\textbf{Sensitive feature Classifier}} \\
\cmidrule(l){4-6}
\textbf{Dataset} & $s$ & metric&\textbf{RF} &\textbf{MLP} &\textbf{XGB} \\ \hline\midrule
\multirow{10}{*}{Adult} &\multirow{5}{*}{gender} &AUC$\uparrow$ &0.9402 &0.9363 &\textbf{0.9413} \\
& &ACC $\uparrow$ &0.8539 &\textbf{0.8559} &0.8463 \\
& &Precision $\uparrow$ &0.9043 &0.9065 &\textbf{0.9549} \\
& &Recall $\uparrow$ &0.8762 &\textbf{0.8768} &0.8107 \\
& &F1 $\uparrow$ &\textbf{0.8900} &0.8914 &0.8769 \\\cmidrule(lr){2-6}
&\multirow{5}{*}{maritalStatus} &AUC$\uparrow$ &0.9883 &0.9882 &\textbf{0.9907} \\
& &ACC $\uparrow$ &\textbf{0.9830} &0.9825 &0.9828 \\
& &Precision $\uparrow$ &\textbf{1.0000} &0.9986 &\textbf{1.0000} \\
& &Recall $\uparrow$ &0.9644 &\textbf{0.9649} &0.9640 \\
& &F1 $\uparrow$ &\textbf{0.9819} &0.9814 &0.9816 \\\midrule
\multirow{10}{*}{Adult-deb} &\multirow{5}{*}{gender} &AUC$\uparrow$ &\textbf{0.8028} &0.8010 &0.7896 \\
& &ACC $\uparrow$ &\textbf{0.7482} &0.7480 &0.7444 \\
& &Precision $\uparrow$ &0.7699 &0.7832 &\textbf{0.8111} \\
& &Recall $\uparrow$ &\textbf{0.8942} &0.8664 &0.8100 \\
& &F1 $\uparrow$ &\textbf{0.8274} &0.8227 &0.8106 \\\cmidrule(lr){2-6}
&\multirow{5}{*}{maritalStatus} &AUC$\uparrow$ &0.7286 &0.7103 &\textbf{0.7708} \\
& &ACC $\uparrow$ &0.6655 &0.6611 &\textbf{0.6918} \\
& &Precision $\uparrow$ &0.6598 &0.6547 &\textbf{0.6677} \\
& &Recall $\uparrow$ &0.6211 &0.6169 &\textbf{0.7098} \\
& &F1 $\uparrow$ &0.6398 &0.6353 &\textbf{0.6879} \\\midrule
\multirow{5}{*}{Crime} &\multirow{5}{*}{race} &AUC$\uparrow$ &0.9893 &0.9885 &\textbf{0.9910} \\
& &ACC $\uparrow$ &0.9450 &\textbf{0.9500} &0.9450 \\
& &Precision $\uparrow$ &0.9412 &\textbf{0.9417} &0.9412 \\
& &Recall $\uparrow$ &0.9655 &\textbf{0.9741} &0.9655 \\
& &F1 $\uparrow$ &0.9532 &\textbf{0.9576} &0.9532 \\\midrule
\multirow{10}{*}{German} &\multirow{5}{*}{gender} &AUC$\uparrow$ &0.7106 &0.5091 &\textbf{0.7139} \\
& &ACC $\uparrow$ &\textbf{0.7300} &0.6900 &0.6900 \\
& &Precision $\uparrow$ &0.7234 &0.6900 &\textbf{0.7879} \\
& &Recall $\uparrow$ &0.9855 &\textbf{1.0000} &0.7536 \\
& &F1 $\uparrow$ &\textbf{0.8344} &0.8166 &0.7704 \\\cmidrule(lr){2-6}
&\multirow{5}{*}{age} &AUC$\uparrow$ &\textbf{0.8876} &0.4756 &0.8363 \\
& &ACC $\uparrow$ &\textbf{0.8600} &0.8100 &0.8100 \\
& &Precision $\uparrow$ &0.8526 &0.8100 &\textbf{0.8605} \\
& &Recall $\uparrow$ &\textbf{1.0000} &\textbf{1.0000} &0.9136 \\
& &F1 $\uparrow$ &\textbf{0.9205} &0.8950 &0.8862 \\
\hline\bottomrule\hline
\end{tabular}
\end{table}

For the sake of reproducibility and reliability, the counterfactuals are generated by a third-party counterfactual framework.
We used DiCE~\cite{mothilal2020dice,10.1145/3351095.3372850}, an open-source framework developed by Microsoft.
DiCE offers several strategies for generating 
counterfactual samples. We used the Genetic and KDtree strategies. The number of counterfactuals generated for each negatively predicted sample is equal to 100 (i.e., $|\mathcal{C}_\mathbf{x}|=100$). We tried to use another agnostic generator, MACE \cite{DBLP:conf/aistats/KarimiBBV20}, but the only compatible models are LR, DT, RF, and MLP (only for a 10-neuron single layer which is not compatible with our MLP-tuned parameters). Moreover, MACE never manage to generate the required number of counterfactuals making the analysis impractical. Some statistics are available in Table~\ref{tab:StatisticsCF}. We can see that MACE is not able to generate the required number of counterfactual samples and, in some cases, there are samples with no counterfactual generated. The bigger the number of features, the more time it requires to generate counterfactual samples. Indeed, as we can see for the DT (Crime) and RF (Adult and Crime), the computation time is extensive and in some cases inestimable (i.e., \textbf{E.C.T.}) while, for MLP, the generation strategy is not compatible since MACE only manage 10-neuron single layer MLP models which are not compatible with our MLP-tuned ones.

\paragraph{Sensitive-Feature Classifier.}
This component plays a crucial role in our methodology since it allows the system to discover hidden discriminatory models.
For the sensitive feature (i.e., gender for Adult, Adult-debiased, and German, and race for Crime), a classifier is thus learned. 
We exploited XGB for implementing this component due to its capability to learn non-linear dependencies.

\paragraph{Metrics.} We evaluate the classifiers' performance with confusion matrix-based metrics (i.e., Accuracy, Recall, Precision, and F1 score) and the Area~Under~the~Receiver~Operative~Curve~(AUC). A preliminary classifiers' fairness evaluation is conducted with the Difference in Statistical Parity\footnote{\scriptsize $\mathrm{DSP} = |\mathds{P}(\hat{Y}=1 \mid S=1) - \mathds{P}(\hat{Y}=1 \mid S=0)|$} (DSP), Difference in Equal Opportunity\footnote{\scriptsize $\mathrm{DEO} = |\mathds{P}(\hat{Y}=1 \mid S=1,Y=1) - \mathds{P}(\hat{Y}=1 \mid S=0,Y=1)|$} (DEO), and Difference in Average Odds\footnote{\scriptsize $\mathrm{DAO} = \frac{1}{2}|\sum_{Y \in \{0,1\}}(\mathds{P}(\hat{Y}=1 \mid S=1,Y) - \mathds{P}(\hat{Y}=1 \mid S=0,Y))$} (DAO).

\paragraph{Split and Hyperparameter Tuning.} The datasets have been split with the hold-out method 90/10 train-test set, with stratified sampling based on the target variable and the sensitive features' labels, to respect original distribution (see Table~\ref{tab:datasetSplit} for split dimension and Table~\ref{tab:SFdataDisrib} for target and sensitive feature distribution in each split). 
We used the Scikit-learn implementation of the train-test split with a random seed set to 42.
The \textit{Decision Maker} and the \textit{Sensitive-feature classifier} models have been tuned on the training set with a Grid Search 5-fold cross-validation methodology, the first optimizing AUC metric, and the latter optimizing F1 score to prevent unbalanced predictions on the sensitive feature.

\subsection{Additional Experiments}\label{sec:AppendixExtendedExperiments}
\begin{figure*}
    \centering
    \includegraphics[width=0.8\textwidth]{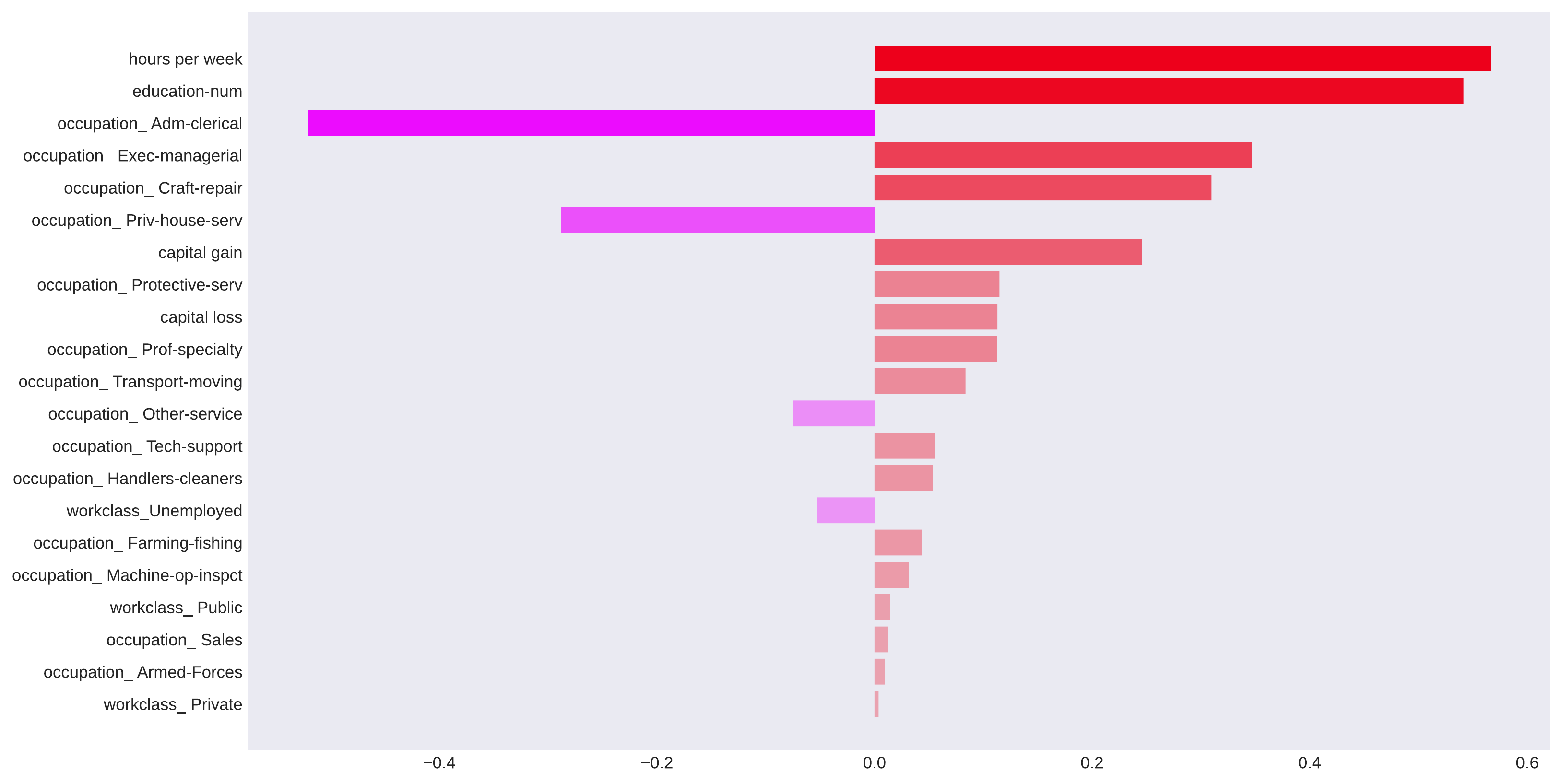}
    \caption{Full list of features correlation with a gender Flip (i.e., $\rho(\epsilon,\delta)$). In the bar chart, counterfactuals were generated by $g(\cdot)$ on MLP \textit{decision-maker} boundary (i.e., $f(\cdot)$), and MLP as \textit{sensitive-feature classifier} (i.e., $f_s(\cdot)$) performed on the Adult-debiased dataset.}
    \label{fig:CorrelationFull}
\end{figure*}

In this section, we can find the extended results of our analysis. Table~\ref{tab:extSFclf} present the full list of metric results for each sensitive feature of the various \textit{Sensitive-Feature Classifiers} (i.e., RF, MLP, XGB). Table~\ref{tab:extCLF} present the full list of accuracy metric results for each \textit{Decision-Maker} and fairness metrics respect each sensitive information. Table~\ref{tab:extDeltaCFLIPS} compare the metric $\Delta\mathrm{CFlips}$ for different \textit{Sensitive-Feature Classifier} and for each sensitive information. Finally, Figure~\ref{fig:CorrelationFull} presents the full list of features correlation with a gender Flip (i.e., $\rho(\mathbf{\epsilon,\delta})$) presented in Figure~\ref{fig:explainCorr}. 

\begin{table*}[!h]\centering
\caption{Complete accuracy and fairness results on the Adult, Adult-debiased, Crime, and
German test set of the Decision Maker Classifiers.}\label{tab:extCLF}
\scriptsize
\renewcommand{\arraystretch}{0.9}
\begin{tabular}{llrrrrrrrrrrr}\hline\toprule
& &\multicolumn{10}{c}{\textbf{Decision-Maker} $f(\cdot)$} \\\cmidrule(lr){3-12}
Dataset &metric &\textbf{LR} &\textbf{DT} &\textbf{SVM} &\textbf{LGBM} &\textbf{XGB} &\textbf{RF} &\textbf{MLP} &\textbf{LFERM} &\textbf{ADV} &\textbf{FairC} \\\hline\midrule
\multirow{11}{*}{Adult} &AUC $\uparrow$ &0.9078 &0.8484 &0.9073 &0.9304 &\textbf{0.9314} &0.9118 &0.9119 &0.9031 &0.9123 &0.8770 \\
&ACC $\uparrow$ &0.8099 &0.8161 &0.8541 &0.8658 &\textbf{0.8698} &0.8534 &0.8494 &0.8428 &0.8512 &0.8395 \\
&Precision $\uparrow$ &0.5782 &0.6879 &0.7570 &0.7655 &\textbf{0.7737} &0.7371 &0.7222 &0.7324 &0.7500 &0.7382 \\
&Recall $\uparrow$ &\textbf{0.8608} &0.4719 &0.6057 &0.6610 &0.6708 &0.6351 &0.6378 &0.5763 &0.5995 &0.5459 \\
&F1 $\uparrow$ &0.6918 &0.5598 &0.6729 &0.7094 &\textbf{0.7186} &0.6823 &0.6774 &0.6450 &0.6663 &0.6277 \\ \cmidrule(lr){2-12}
&DSP $\downarrow$ (\textit{gender}) &0.2947 &0.1461 &0.1769 &0.1850 &0.1884 &0.1854 &0.1902 &0.1448 &0.1151 &\textbf{0.0528} \\
&DEO $\downarrow$ (\textit{gender}) &0.0546 &0.0760 &0.0644 &0.0379 &0.0635 &0.0216 &0.0529 &\textbf{0.0194} &0.1399 &0.2451 \\
&DAO $\downarrow$ (\textit{gender}) &0.1241 &0.0722 &0.0692 &0.0569 &0.0680 &0.0545 &0.0708 &\textbf{0.0386} &0.0879 &0.1274 \\ \cmidrule(lr){2-12}
&DSP $\downarrow$ (\textit{maritalStatus}) &0.6290 &0.2833 &0.3562 &0.3648 &0.3601 &0.3750 &0.3430 &0.2713 &0.3779 &\textbf{0.0162} \\
&DEO $\downarrow$ (\textit{maritalStatus}) &0.4656 &0.3461 &0.3464 &0.2893 &0.2874 &0.3014 &0.3168 &\textbf{0.1875} &0.3851 &0.3710 \\
&DAO $\downarrow$ (\textit{maritalStatus}) &0.4736 &0.2343 &0.2554 &0.2148 &0.2122 &0.2388 &0.2300 &\textbf{0.1467} &0.2859 &0.2112 \\\midrule
\multirow{11}{*}{Adult-deb} &AUC $\uparrow$ &0.8233 &0.7895 &0.7944 &\textbf{0.8596} &0.8578 &0.8336 &0.8271 &0.8017 &0.8309 &0.7981 \\
&ACC $\uparrow$ &0.7367 &0.8017 &0.8061 &0.8371 &\textbf{0.8375} &0.8267 &0.8156 &0.7953 &0.8196 &0.8054 \\
&Precision $\uparrow$ &0.4790 &0.8294 &\textbf{0.8389} &0.8038 &0.8063 &0.7621 &0.7540 &0.7079 &0.7529 &0.7526 \\
&Recall $\uparrow$ &\textbf{0.7119} &0.2516 &0.2694 &0.4532 &0.4532 &0.4371 &0.3800 &0.2962 &0.4050 &0.3202 \\
&F1 $\uparrow$ &0.5727 &0.3860 &0.4078 &0.5796 &\textbf{0.5802} &0.5556 &0.5053 &0.4176 &0.5267 &0.4493 \\ \cmidrule(lr){2-12}
&DSP $\downarrow$ (\textit{gender}) &0.1567 &\textbf{0.0438} &0.0534 &0.1093 &0.1056 &0.1058 &0.0863 &0.0639 &0.0957 &0.0575 \\
&DEO $\downarrow$ (\textit{gender}) &0.0695 &0.0492 &0.0353 &0.0470 &0.0400 &0.0703 &\textbf{0.0173} &0.0179 &0.0326 &0.0529 \\
&DAO $\downarrow$ (\textit{gender}) &0.0693 &0.0272 &0.0227 &0.0356 &0.0304 &0.0461 &0.0188 &\textbf{0.0186} &0.0282 &0.0315 \\ \cmidrule(lr){2-12}
&DSP $\downarrow$ (\textit{maritalStatus}) &0.1793 &0.0945 &0.0948 &0.1702 &0.1663 &0.1501 &0.1249 &\textbf{0.0336} &0.1316 &0.1241 \\
&DEO $\downarrow$ (\textit{maritalStatus}) &0.1450 &0.0468 &0.0720 &0.0676 &0.0645 &\textbf{0.0460} &0.1266 &0.0489 &0.1108 &0.1128 \\
&DAO $\downarrow$ (\textit{maritalStatus}) &0.0880 &0.0240 &0.0362 &0.0409 &0.0355 &\textbf{0.0232} &0.0633 &0.0262 &0.0591 &0.0570 \\\midrule
\multirow{8}{*}{Crime} &AUC $\uparrow$ &0.9248 &0.8991 &\textbf{0.9288} &0.9168 &0.9099 &0.9096 &0.9203 &0.9100 &0.9008 &0.8024 \\
&ACC $\uparrow$ &\textbf{0.8700} &0.8200 &\textbf{0.8700} &0.8400 &0.8500 &0.8400 &0.8650 &0.8400 &0.8100 &0.7500 \\
&Precision $\uparrow$ &0.8627 &0.8265 &\textbf{0.8776} &0.8400 &0.8500 &0.8400 &0.8544 &0.8333 &0.8444 &0.7500 \\
&Recall $\uparrow$ &\textbf{0.8800} &0.8100 &0.8600 &0.8400 &0.8500 &0.8400 &\textbf{0.8800} &0.8500 &0.7600 &0.7500 \\
&F1 $\uparrow$ &\textbf{0.8713} &0.8182 &0.8687 &0.8400 &0.8500 &0.8400 &0.8670 &0.8416 &0.8000 &0.7500 \\ \cmidrule(lr){2-12}
&DSP $\downarrow$ (\textit{race}) &0.6535 &0.6190 &0.6396 &0.6363 &0.6568 &0.6363 &0.6622 &0.6125 &0.5501 &\textbf{0.2258} \\
&DEO $\downarrow$ (\textit{race}) &0.3294 &0.4039 &0.3843 &0.2824 &0.2941 &0.2824 &0.3294 &0.2941 &0.1882 &\textbf{0.1373} \\
&DAO $\downarrow$ (\textit{race}) &0.3438 &0.3827 &0.3390 &0.3525 &0.3656 &0.3525 &0.3599 &0.3278 &0.2732 &\textbf{0.0862} \\\midrule
\multirow{11}{*}{German} &AUC $\uparrow$ &\textbf{0.8186} &0.7219 &0.8110 &0.7614 &0.7871 &0.7936 &0.8162 &0.7605 &0.7371 &0.8152 \\
&ACC $\uparrow$ &0.7600 &0.7600 &0.7600 &0.7500 &\textbf{0.7900} &0.7600 &0.7600 &0.7200 &0.7300 &0.7400 \\
&Precision $\uparrow$ &\textbf{0.8485} &0.7805 &0.7738 &0.7848 &0.8025 &0.7674 &0.7738 &0.7188 &0.7792 &0.7619 \\
&Recall $\uparrow$ &0.8000 &0.9143 &0.9286 &0.8857 &0.9286 &0.9429 &0.9286 &\textbf{0.9857} &0.8571 &0.9143 \\
&F1 $\uparrow$ &0.8235 &0.8421 &0.8442 &0.8322 &\textbf{0.8609} &0.8462 &0.8442 &0.8313 &0.8163 &0.8312 \\ \cmidrule(lr){2-12}
&DSP $\downarrow$ (\textit{gender}) &0.1187 &\textbf{0.0271} &0.0449 &0.1632 &0.0519 &0.0626 &0.0449 &0.0355 &0.1809 &0.0449 \\
&DEO $\downarrow$ (\textit{gender}) &0.1400 &0.0500 &0.0300 &0.1900 &0.0400 &0.0800 &0.0300 &\textbf{0.0200} &0.2200 &0.0500 \\
&DAO $\downarrow$ (\textit{gender}) &0.1657 &0.0537 &0.0892 &0.1117 &\textbf{0.0296} &0.0878 &0.0892 &0.0746 &0.1267 &0.0728 \\ \cmidrule(lr){2-12}
&DSP $\downarrow$ (\textit{age}) &0.2827 &0.0845 &0.0344 &0.1676 &0.0006 &0.1397 &0.2112 &\textbf{0.0000} &0.0273 &0.1956 \\
&DEO $\downarrow$ (\textit{age}) &0.3020 &0.0693 &0.0570 &0.0740 &0.0847 &0.0570 &0.2049 &\textbf{0.0000} &0.0108 &0.2049 \\
&DAO $\downarrow$ (\textit{age}) &0.1851 &0.0347 &0.0853 &0.0995 &0.0651 &0.0967 &0.1195 &\textbf{0.0000} &0.0849 &0.1252 \\
\hline\bottomrule\hline
\end{tabular}
\end{table*}

\begin{table}\centering
\caption{Full Table of $\Delta$CFlips (\%) metrics with different $f_s(\cdot)$ (i.e., RF, MLP, and XGB) on the Adult, Adult-debiased, Crime, and German test set. We mark the best-performing method for each metric in bold font. For the German dataset, the MLP is not able to predict sample $\mathbf{x}$ sensitive information resulting in no sample available for the CFlips analysis.}\label{tab:extDeltaCFLIPS}
\scriptsize
\setlength{\tabcolsep}{2.5pt}
\begin{tabular}{lllrr|rr|rr}\hline\toprule
& & &\multicolumn{6}{c}{$\Delta\mathrm{CFlips}$} \\ \cmidrule(lr){4-9}
& & &\multicolumn{2}{c}{\textbf{RF}} &\multicolumn{2}{c}{\textbf{MLP}} &\multicolumn{2}{c}{\textbf{XGB}} \\\cmidrule(lr){4-5} \cmidrule(lr){6-7} \cmidrule(lr){8-9}
Dataset &s &$f(\cdot)$ &Genetic &\multicolumn{1}{c}{KDtree} &Genetic &\multicolumn{1}{c}{KDtree} &Genetic &\multicolumn{1}{c}{KDtree} \\\hline\midrule
\multirow{20}{*}{Adult} &\multirow{10}{*}{\textit{gender}} &LR &67.35 &76.09 &66.75 &75.28 &67.05 &76.03 \\ 
& &DT &70.10 &77.16 &70.07 &77.14 &70.23 &77.14 \\
& &SVM &74.03 &79.33 &73.99 &79.82 &73.99 &79.35 \\
& &LGBM &70.72 &78.40 &70.26 &77.57 &70.87 &78.40 \\
& &XGB &70.34 &78.59 &69.54 &77.42 &70.40 &78.42 \\
& &RF &73.24 &77.01 &72.25 &75.68 &73.09 &76.68 \\
& &MLP &71.17 &77.86 &71.24 &77.80 &71.32 &77.95 \\
& &LFERM &32.15 &68.13 &32.33 &67.86 &32.40 &68.25 \\
& &ADV &\textbf{4.36} &53.70 &\textbf{4.25} &53.80 &\textbf{7.96} &53.90 \\
& &FairC &43.09 &\textbf{35.86} &43.51 &\textbf{35.78} &38.72 &\textbf{36.27} \\\cmidrule(lr){2-9}
&\multirow{10}{*}{\textit{maritalStatus}} &LR &96.08 &93.30 &96.36 &93.22 &96.36 &93.27 \\
& &DT &82.95 &89.41 &82.63 &89.15 &82.60 &89.13 \\
& &SVM &96.08 &96.13 &96.16 &96.06 &96.24 &96.06 \\
& &LGBM &94.94 &94.33 &94.85 &94.25 &94.90 &94.27 \\
& &XGB &94.99 &94.35 &94.93 &94.25 &94.94 &94.27 \\
& &RF &93.45 &90.26 &93.39 &90.14 &93.41 &90.17 \\
& &MLP &95.34 &95.26 &95.37 &95.23 &95.36 &95.25 \\
& &LFERM &86.96 &88.71 &87.03 &88.59 &87.04 &88.60 \\
& &ADV &\textbf{15.98} &\textbf{32.96} &\textbf{15.38} &\textbf{33.39} &\textbf{15.42} &\textbf{33.40} \\ 
& &FairC &32.06 &97.52 &54.77 &97.47 &38.77 &97.48 \\
\midrule
\multirow{20}{*}{Adult-deb} &\multirow{10}{*}{\textit{gender}} &LR &85.74 &84.17 &82.35 &72.67 &40.02 &\textbf{8.37} \\
& &DT &75.87 &99.19 &71.11 &91.46 &39.65 &12.74 \\
& &SVM &66.67 &77.20 &86.17 &\textbf{54.70} &\textbf{18.27} &\textbf{11.66} \\
& &LGBM &89.65 &97.25 &92.22 &95.68 &56.16 &59.11 \\
& &XGB &96.04 &98.34 &93.11 &95.21 &69.02 &59.22 \\
& &RF &97.85 &95.30 &92.38 &93.37 &71.12 &83.87 \\
& &MLP &99.50 &98.68 &99.48 &98.47 &92.38 &89.37 \\
& &LFERM &\textbf{54.41} &\textbf{70.05} &\textbf{51.67} &70.00 &39.67 &60.55 \\
& &ADV &69.94 &97.48 &56.37 &96.92 &33.17 &85.81 \\
& &FairC &89.29 &96.18 &90.77 &97.89 &80.12 &72.24 \\\cmidrule(lr){2-9}
&\multirow{10}{*}{\textit{maritalStatus}} &LR &34.15 &28.55 &46.32 &7.10 &6.55 &25.75 \\
& &DT &94.60 &84.99 &74.97 &48.76 &87.69 &69.05 \\
& &SVM &85.83 &24.97 &93.58 &\textbf{0.76} &\textbf{1.94} &\textbf{19.41} \\
& &LGBM &89.86 &92.08 &62.65 &53.42 &59.29 &91.66 \\
& &XGB &88.51 &91.90 &68.05 &56.31 &43.23 &89.76 \\
& &RF &79.39 &89.18 &57.41 &79.50 &70.60 &71.01 \\
& &MLP &96.04 &82.57 &91.13 &95.82 &85.62 &52.85 \\
& &LFERM &20.47 &\textbf{16.79} &48.09 &67.22 &31.58 &34.28 \\
& &ADV &29.38 &78.39 &14.51 &92.83 &24.69 &43.80 \\
& &FairC &\textbf{0.00} &84.38 &\textbf{0.00} &96.22 &\textbf{0.00} &63.25 \\\midrule
\multirow{10}{*}{Crime} &\multirow{10}{*}{\textit{race}} &LR &80.56 &78.33 &82.57 &86.31 &81.70 &78.76 \\
& &DT &72.48 &66.04 &74.13 &71.76 &72.80 &66.25 \\
& &SVM &79.02 &75.31 &83.60 &83.36 &79.17 &75.15 \\
& &LGBM &78.23 &76.65 &82.07 &85.11 &78.22 &76.52 \\
& &XGB &78.93 &76.84 &81.46 &85.35 &78.88 &76.79 \\
& &RF &70.27 &68.12 &72.87 &73.84 &70.23 &68.19 \\
& &MLP &76.90 &74.66 &82.32 &82.31 &76.41 &75.22 \\
& &LFERM &68.42 &63.85 &71.16 &71.44 &68.68 &63.69 \\
& &ADV &74.51 &74.10 &74.42 &81.80 &74.09 &74.12 \\
& &FairC &\textbf{25.50} &\textbf{17.60} &\textbf{31.77} &\textbf{25.13} &\textbf{27.06} &\textbf{18.40} \\\midrule
\multirow{20}{*}{German} &\multirow{10}{*}{\textit{gender}} &LR &39.18 &77.78 &0.00 &0.00 &11.89 &35.65 \\
& &DT &7.30 &8.31 &0.00 &0.00 &21.61 &22.43 \\
& &SVM &5.89 &8.00 &0.00 &0.00 &38.64 &37.29 \\
& &LGBM &78.09 &71.72 &0.00 &0.00 &\textbf{3.54} &28.54 \\
& &XGB &17.17 &70.89 &0.00 &0.00 &7.85 &32.37 \\
& &RF &\textbf{4.24} &7.35 &0.00 &0.00 &15.44 &37.17 \\
& &MLP &5.33 &8.67 &0.00 &0.00 &29.07 &29.00 \\
& &LFERM &6.57 &\textbf{3.03} &0.00 &0.00 &24.24 &23.23 \\
& &ADV &82.32 &70.92 &0.00 &0.00 &18.46 &\textbf{31.51} \\
& &FairC &5.22 &10.02 &0.00 &0.00 &23.94 &30.81 \\\cmidrule(lr){2-9}
&\multirow{10}{*}{age} &LR &80.66 &78.28 &0.00 &0.00 &73.63 &62.61 \\
& &DT &78.48 &68.91 &0.00 &0.00 &66.85 &58.82 \\
& &SVM &4.49 &9.49 &0.00 &0.00 &11.11 &16.02 \\
& &LGBM &68.10 &68.52 &0.00 &0.00 &70.48 &60.94 \\
& &XGB &3.27 &7.43 &0.00 &0.00 &8.30 &14.43 \\
& &RF &3.17 &8.79 &0.00 &0.00 &8.59 &15.15 \\
& &MLP &73.17 &65.42 &0.00 &0.00 &58.08 &53.75 \\
& &LFERM &\textbf{0.00} &\textbf{0.00} &0.00 &0.00 &\textbf{0.00} &\textbf{0.0} \\
& &ADV &85.35 &80.23 &0.00 &0.00 &68.44 &61.70 \\
& &FairC &82.15 &66.80 &0.00 &0.00 &60.06 &48.41 \\
\hline\bottomrule\hline
\end{tabular}
\end{table}

\newpage
\clearpage
\newpage

\end{document}